\documentclass{article}

\PassOptionsToPackage{numbers,compress}{natbib}

\usepackage[eandd,final]{neurips_2026}

\usepackage[utf8]{inputenc}
\usepackage[T1]{fontenc}
\usepackage[hidelinks]{hyperref}
\usepackage{url}
\usepackage{booktabs}
\usepackage{amsfonts}
\usepackage{amsmath}
\usepackage{amssymb}
\usepackage{nicefrac}
\usepackage{microtype}
\usepackage[table]{xcolor}

\newif\ifcamerareadyedits
\camerareadyeditsfalse
\DeclareRobustCommand{\cradded}[1]{%
  \ifcamerareadyedits{\color{red}#1}\else#1\fi}
\newenvironment{crrevision}{%
  \begingroup\ifcamerareadyedits\color{red}\fi}{\endgroup}
\expandafter\def\csname crbib@11125616\endcsname{}
\expandafter\def\csname crbib@hong2025motionbench\endcsname{}
\expandafter\def\csname crbib@motamed2026physicsiq\endcsname{}
\expandafter\def\csname crbib@zheng2025vbench2\endcsname{}
\expandafter\def\csname crbib@openai2026gpt56sol\endcsname{}
\expandafter\def\csname crbib@google2026gemini31pro\endcsname{}

\usepackage{graphicx}
\usepackage{multirow}
\usepackage{longtable}
\usepackage{wrapfig}
\usepackage{tcolorbox}
\usepackage{soul}
\usepackage{bbm}
\usepackage{tikz}
\usetikzlibrary{calc}
\usepackage{xspace}
\usepackage{caption}

\definecolor{cvprblue}{rgb}{0.21,0.49,0.74}
\definecolor{row_color}{HTML}{F3F8FC}
\definecolor{QuantiDarkYellow}{RGB}{241,223,186}
\definecolor{QuantiLightYellow}{RGB}{248,236,208}
\definecolor{QuantiThirdYellow}{RGB}{252,243,224}
\newcommand{\best}[1]{\cellcolor{QuantiDarkYellow}\textbf{#1}}
\newcommand{\secondbest}[1]{\cellcolor{QuantiLightYellow}\textbf{#1}}
\newcommand{\thirdbest}[1]{\cellcolor{QuantiThirdYellow}\textbf{#1}}

\makeatletter
\newcommand{\printfnsymbol}[1]{\textsuperscript{\@fnsymbol{#1}}}
\makeatother

\title{HumanScore: Benchmarking Human Motions in Generated Videos}

\author{%
  \cradded{Tiange Xiang$^{1*}$ \quad Yusu Fang$^{1,2*}$ \quad Tian Tan$^1$ \quad Narayan Schuetz$^1$}\\
  \cradded{\textbf{Scott Delp$^1$ \quad Li Fei-Fei$^{1\dagger}$ \quad Ehsan Adeli$^{1\dagger}$}}\\
  \cradded{$^1$Stanford University \quad $^2$Peking University}\\
  \cradded{\small $^*$Equal first authorship \quad $^\dagger$Equal last authorship}
}

\begin{document}

\maketitle

\begin{center}
  \vspace{-2.5em}
  \cradded{\small\url{https://cs.stanford.edu/~xtiange/projects/humanscore/}}
\end{center}

\begin{center}
    \vspace{-1em}
    \includegraphics[width=0.99\textwidth]{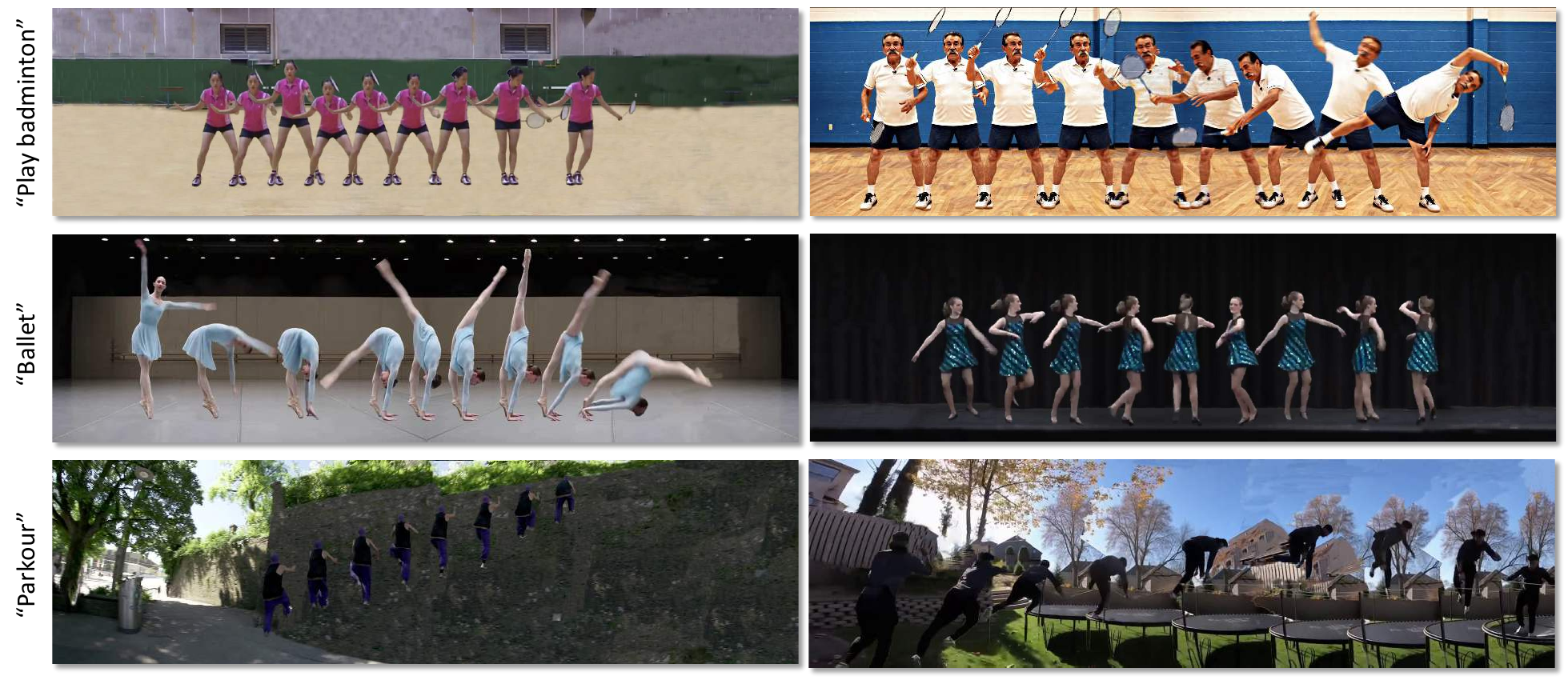}
    \captionsetup{hypcap=false}
    \captionof{figure}{\textbf{Real or AI?} This figure compares frames from real videos alongside AI-generated ones from modern video generators. Can you tell which is which? Processed full videos are in supplementary material and answers are in the footnote\protect\footnotemark. \textbf{HumanScore} is designed to detect subtle biomechanical violations produced by AI video generators, even when they are imperceptible to the human eye.}
    \label{fig:teaser}
\end{center}
\footnotetext{[Spoiler Alert] We recommend viewing the full videos in the supplementary material first. The AI-generated videos are: Play badminton (right), Ballet (left), and Parkour (right).}

\begin{abstract}
Recent advances in model architectures, compute, and data scale have driven rapid progress in video generation, producing increasingly realistic content. Yet standard video benchmarks still provide limited diagnostic insight into whether generated people move in anatomically consistent, kinematically feasible, and temporally plausible ways. We present \textbf{HumanScore}, a controlled benchmark for evaluating single-person, full-body human motion quality in AI-generated videos. HumanScore combines a curated suite of $51$ motion types with gentle and intense variants, yielding $102$ standardized prompts, and six interpretable metrics spanning anatomical correctness, kinematic plausibility, and \cradded{kinetic plausibility}. Because \cradded{several} metrics operate on reconstructed 3D motion from monocular video, HumanScore is designed as an estimator-aware diagnostic protocol rather than a perfect biomechanical oracle, using tolerance margins, robust aggregation, and robustness analyses to support interpretation. Evaluating thirteen state-of-the-art open-source and proprietary video generators, we find that visually compelling generations still exhibit recurrent motion-level failures such as temporal jitter, anatomically implausible poses, and motion drift.

\end{abstract}

\section{Introduction}
\label{sec:intro}

Modern video generators trained on large-scale data produce compelling scenes with realistic appearance and score highly on benchmarks such as VBench~\cite{10657096} and T2V-CompBench~\cite{11092317}. Yet these benchmarks offer limited diagnostic evidence about whether generated \emph{people} move in physically plausible ways: human-centric clips can exhibit duplicated limbs, unstable body proportions, implausible joint configurations, velocity spikes, and temporal jitter even when each frame looks photorealistic. We argue that controlled, biomechanics-grounded evaluation of human motion should sit at the core of any benchmark that tracks this dimension of realism.

This gap matters because human-centric content is a dominant use case for video generation across entertainment, education, sports, and telepresence---domains where credibility and safety depend not only on how things \emph{look}, but on how plausibly people \emph{move}. Existing metrics that emphasize pixel realism, semantic alignment, or optical-flow smoothness leave motion realism under-investigated, offering little insight into where and why a clip fails.

\begin{wrapfigure}{r}{0.46\textwidth}
    \vspace{-1.3em}
    \centering
    \includegraphics[width=0.9\linewidth]{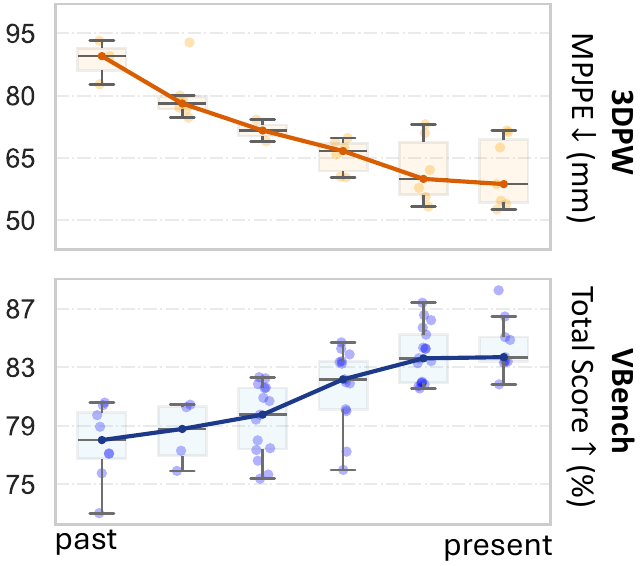}
    \caption{Human pose estimators are converging at low errors, while video generators are becoming more realistic.}
    \label{fig:hmr_results}
    \vspace{-1em}
\end{wrapfigure}

Evaluating motion realism requires moving beyond pixels to 3D structure and dynamics that capture biomechanics, kinematics, and kinetics (e.g., limb-length consistency, range of motion, acceleration feasibility). Doing this from monocular video has long been hard, but two trends now make it tractable (\figureautorefname~\ref{fig:hmr_results}): human pose and mesh recovery have saturated on benchmarks such as 3DPW~\cite{von2018recovering}, signaling estimators accurate enough to support analysis; concurrently, video generators are saturating standard visual benchmarks. This convergence---stable pose estimators meeting visually saturated generators---makes it timely and necessary to push for principled, quantitative evaluation of motion realism. See \figureautorefname~\ref{fig:teaser} and try to tell real videos apart from those produced by today's generators.

We introduce a benchmark focused on \emph{controlled single-person, full-body human motion realism} in generated videos. \emph{HumanScore does not propose a new pose estimator; it repurposes mature reconstruction tools into an estimator-aware biomechanical evaluation protocol with confidence filtering, tolerance margins, and robust temporal aggregation.} The benchmark isolates human motion from confounders such as camera motion, severe occlusion, and multi-person interaction, \cradded{to reduce scene-related reconstruction errors when comparing generators}; the resulting evaluation is complementary to appearance-centric metrics and surfaces failure modes that traditional measures do not localize.

Our contributions are three-fold: \textbf{(i)} We curate a controlled prompt suite for single-person, full-body motion generation, covering $51$ motion types across three difficulty levels with gentle and intense variants. \textbf{(ii)} We define six interpretable, estimator-aware metrics grounded in anatomy, kinematics, and kinetics, together with a video-level aggregation that captures violation frequency, severity, and persistence. \textbf{(iii)} We apply HumanScore to thirteen modern video generators and a real-video reference set, revealing broad performance tiers and complementary motion-level failure profiles that visual appearance alone does not capture.

\section{Related Work}
\label{sec:relatedwork}

\subsection{Video Generation Benchmarks}

Recent benchmarks evaluate video generation along a broad spectrum, including visual fidelity, temporal consistency~\cite{kim2024stream}, text alignment, and compositionality. Representative efforts include VBench~\cite{10657096} and its extension VBench++~\cite{huang2024vbench++}, which provide multi-dimensional suites with human-alignment analyses; EvalCrafter~\cite{liu2024evalcrafter}, which aggregates large prompt sets and objective metrics with human correlation studies for better video quality evaluations; T2V-CompBench~\cite{11092317}, which evaluates generated content by targeting compositional generalization across attributes, spatial relations, and action binding; Video-Bench~\cite{11094238}, which presents a toolkit to better cover action consistency and motion/temporal quality; WorldScore~\cite{duan2025worldscore}, emphasizes holistic ``world generation'' quality; HumanBench and MotionBench~\cite{tang2023humanbench,hong2025motionbench}, which focus on human-centric perception or motion understanding. \cradded{VBench-2.0~\cite{zheng2025vbench2} further evaluates intrinsic faithfulness, including human fidelity and physics, using both multimodal evaluators and specialist detectors. Physics-IQ~\cite{motamed2026physicsiq} tests physical understanding through real-world outcome prediction, sharing our premise that visual realism does not imply physical plausibility. HumanScore complements these broader benchmarks with explicit anatomical, joint-geometry, and temporal constraints on reconstructed human motion.} Our benchmark fills this gap by proposing novel and comprehensive metrics for diagnosing correctness and consistency in anatomy, kinematics, and kinetics.

\subsection{Human Motion Evaluation Methods}

Motion-focused evaluation in videos often emphasizes perceptual alignment or distributional similarity rather than explicit biomechanical constraints. As also seen in Video-Bench~\cite{11094238}, such metrics include action consistency, motion quality, and temporal consistency; perception-driven motion metrics (PMM) from VMBench~\cite{ling2025vmbench,ling2025vmbenchbenchmarkperceptionalignedvideo}; recognizer-based Action-Score~\cite{wang2025recognizing,lin2024evaluating,li2024genaibench,camerabench}; and distributional distances such as FVD~\cite{unterthiner2019fvd} and motion-focused variants like FVMD/FMD~\cite{liu2024fr,maiorca2022evaluatingqualitysynthesizedmotion, maiorca2023objective}. These measures are effective for large-scale ranking, yet they do not rationalize or diagnose the unfaithfulness and surrealism of a motion (e.g., when a specific pose is `inhuman').

A parallel line of work evaluates \emph{4D motion generation} in parametric 3D human-body models (e.g., SMPL/SMPL-X~\cite{SMPL:2015,SMPL-X:2019}), with recent unified suites and widely used datasets/protocols~\cite{lin2025quest,Guo_2022_CVPR,Plappert_2016,lin2023motionx,mahmood2019amass}. While commonly adopted in the research community, the target models are specifically motion generation models originally trained on purely real-world motion data. As a result, the generated motions from these models always interpolate between natural motions and therefore rarely violate biomechanics. Moreover, the human shape models used by these motion generation models are typical in their shape and scale, further constraining the variability of generated motions. 

In contrast, our benchmark takes a novel angle by evaluating human motions extracted directly \emph{from AI-generated videos}, which avoids over-constraining assumptions and enables directly translating insights from reconstruction/physics-aware works such as MoYo~\cite{tripathi2023ipman}, NeuHMR~\cite{11125616}, monocular contact reasoning~\cite{10.1007/978-3-030-58558-7_5}, 4D-Humans~\cite{10378229}, PhysPT~\cite{Zhang_2024_CVPR}, MultiPhys~\cite{ugrinovic2024multiphys} and PhysCap~\cite{PhysCapTOG2020} into \emph{mechanism-driven and interpretable} metrics. Our proposed metrics do not contradict existing perceptual- and distribution-based measurements; instead, they serve as complementary tools that specifically target the biomechanical fidelity of generated human motions.

\begin{figure*}[t]
    \centering
    \includegraphics[width=0.99\linewidth]{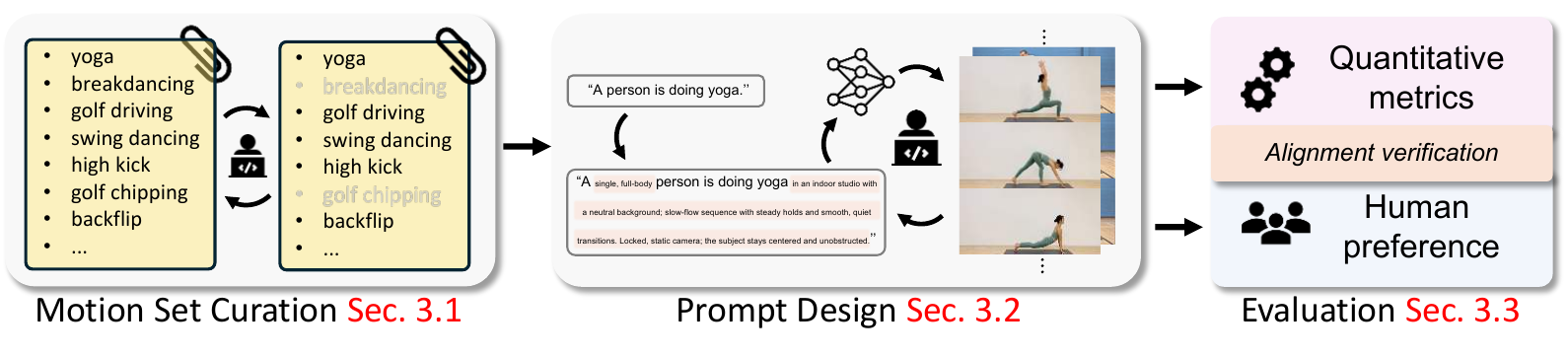}
    \vspace{-1em}
    \caption{\textbf{Overview of HumanScore}. The pipeline begins with curating a representative set of human motions from a large pool of common actions. For each motion, we carefully design prompts to mitigate model-specific biases and ensure consistent conditioning across generators. The refined prompts are then passed to both proprietary and open-source state-of-the-art video generation models. Human verification is incorporated at all stages for quality check. For each evaluation dimension shown in \figureautorefname~\ref{fig:structure}, we design biomechanics-informed quantitative metrics, together with human preference studies to provide comprehensive insights from multiple perspectives.}
    \vspace{-0.5em}
    \label{fig:pipeline}
\end{figure*}

\section{HumanScore}
\label{sec:benchmark}

Evaluating human motion dynamics systematically entails a sequence of components. As outlined in \figureautorefname~\ref{fig:pipeline}, HumanScore contains three key components: (i) curation of the list of motions to benchmark; (ii) carefully designed prompts that elicit reliable, articulated generations without artifacts and ambiguousity; and (iii) evaluation of the generated videos with our proposed multifaceted metrics.

\begin{wrapfigure}{r}{0.43\textwidth}
    \centering
    \vspace{-1.0em}
    \includegraphics[width=0.8\linewidth]{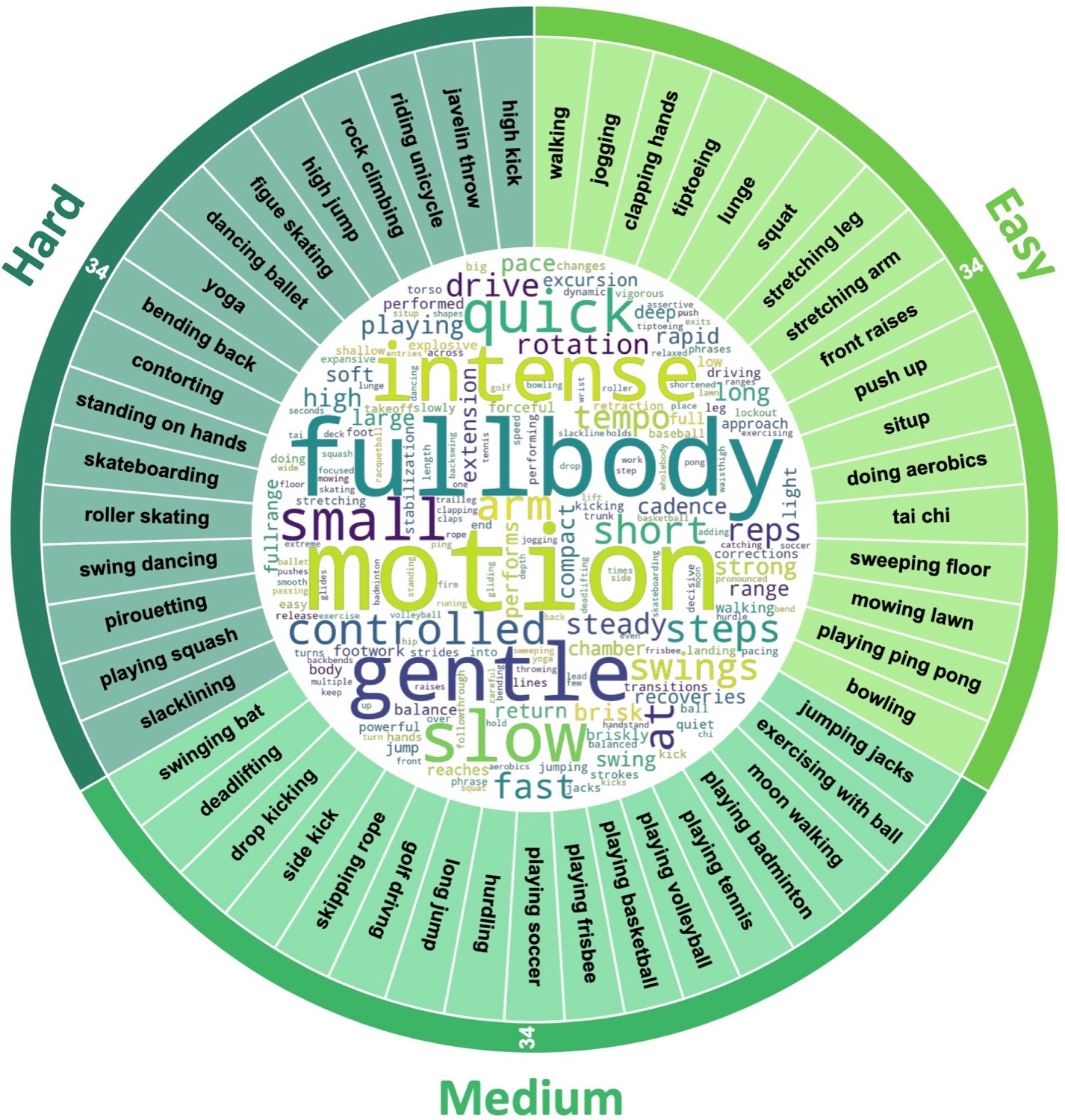}
    \caption{Our curated motion set includes 17 distinct motions for each difficulty level, each with two levels of intensity, resulting in $102$ unique prompts in total. High frequent words across prompts are illustrated.}
    \label{fig:prompt_stats}
    \vspace{-1.3em}
\end{wrapfigure}

\subsection{Motion Set Curation} \label{sec:motion_curation}
The construction of HumanScore begins with a curated set of motions. Because human movement is unstructured and non-deterministic, we instruct video generators with a standardized, well-defined vocabulary of common motion categories to avoid prompt ambiguity.

Since real-world movement cannot be exhaustively enumerated, our goal is a compact but diverse reference pool. We start from Kinetics-700~\cite{carreira2019short}, which is widely used in motion research; because its 700 categories contain semantic redundancies (e.g., `golf driving' vs.\ `golf chipping'), we conduct a sifting process as follows.

Manually selecting a subset from 700 motions requires kinematics expertise and is prohibitively time-consuming. We instead apply a semi-automatic process with three guiding principles followed by an empirical feasibility check. First, we de-duplicate motion types by encoding their names with CLIP~\cite{radford2021learning} and SBERT~\cite{reimers2019sentence} and applying farthest-point sampling on the cosine similarity matrix. Second, we balance categories across motion families using a Large Language Model (LLM) that groups candidates into families such as \textit{Gait and Footwork} (e.g., walking, jogging) or \textit{Gymnastics and Calisthenics} (e.g., ballet, yoga). Third, we prompt the LLM to verify broad coverage across upper-body--dominant, lower-body--dominant, full-body coordination, flips/rotations, self-contact, and object interactions, narrowing the pool from 700 to 120 candidates. Finally, an empirical feasibility check---combining our objective metrics with manual inspection---removes candidates that prove too abstract or consistently fail.
The final result is a curated set of 51 distinct motion types, balanced across three difficulty levels: simple, medium, and hard. Each motion is further provided in two intensity levels (gentle/intense), yielding 102 motion clips in total.

\begin{figure*}[t]
    \centering
    \includegraphics[width=0.99\linewidth]{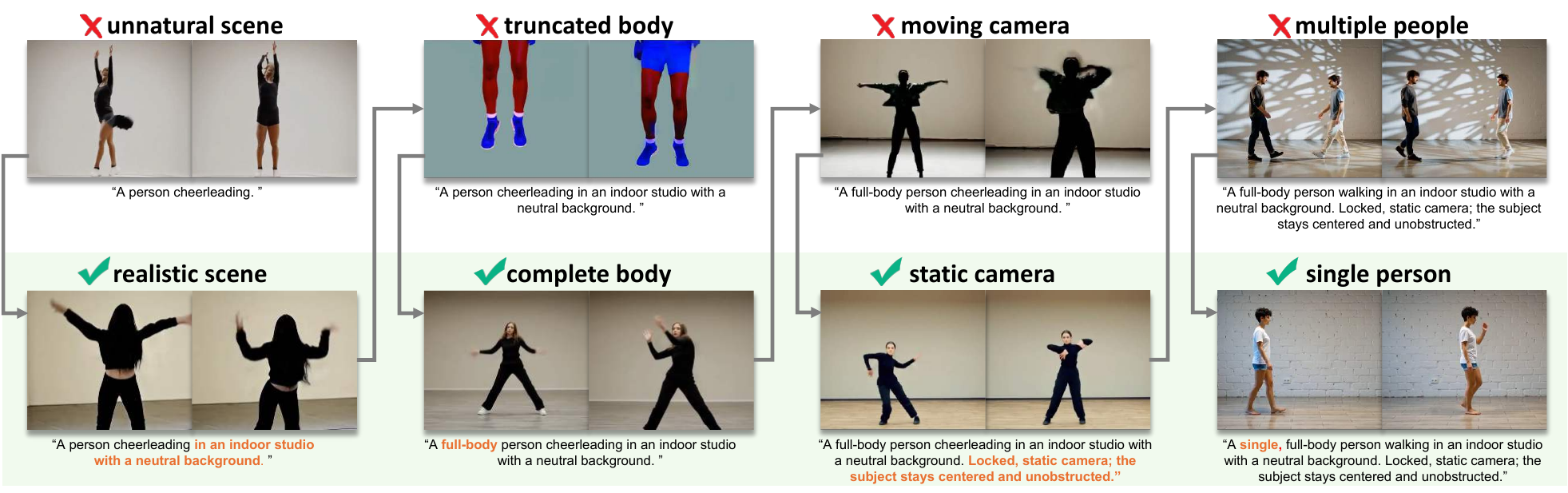}
     \vspace{-0.5em}
    \caption{\textbf{Prompt Design.} Different models tend to have different biases when generating videos, which may lead to unnatural scenes, truncated bodies, moving cameras, or multiple people. We experimented extensively with prompt engineering to mitigate these biases and obtain stable motions in the generated videos, which facilitates metric calculation. Best viewed when zoomed in.}
    \label{fig:prompt_design}
    \vspace{-0.5em}
\end{figure*}

\subsection{Prompt Design} \label{sec:prompt_design}
A benchmark for human motion must be built on precise, unambiguous prompts. Ambiguous prompts (e.g., ``A person cheerleading'') often produce blurred or under-articulated motions unsuitable for evaluation, especially in generators with limited prompt-following capabilities.

We therefore propose a systematic approach to prompt engineering, iteratively refining naive prompts by inspecting their generations to validate each change (\figureautorefname~\ref{fig:prompt_design}). Specifically, to \textbf{(i)} make the motion clearly visible against a non-distracting background, we add \textit{`in an indoor studio with a neutral background'}; to \textbf{(ii)} ensure the complete body is visible for analysis, we specify \textit{`full-body'}; to \textbf{(iii)} decouple subject motion from camera motion, we append \textit{`Locked, static camera; the subject stays centered and unobstructed.'}; and finally, to \textbf{(iv)} avoid multiple subjects, we enforce \textit{`A single person'}.

Combining these standardized elements, our final prompts entail five components---\textbf{Scene}, \textbf{Motion}, \textbf{Intensity}, \textbf{Description}, and \textbf{Camera}---with the same prompt applied across all models for fairness. \cradded{This constrained format reduces scene- and camera-level confounds for controlled cross-model comparison, but limits the scope of the resulting leaderboard.} The metrics themselves are not tied to this format and apply to in-the-wild monocular videos with standard preprocessing (see supplementary materials for details).

\subsection{Metric Design} \label{sec:metric_design}

\begin{wrapfigure}{r}{0.58\textwidth}
    \vspace{0.6em}
    \centering
    \includegraphics[width=0.9\linewidth]{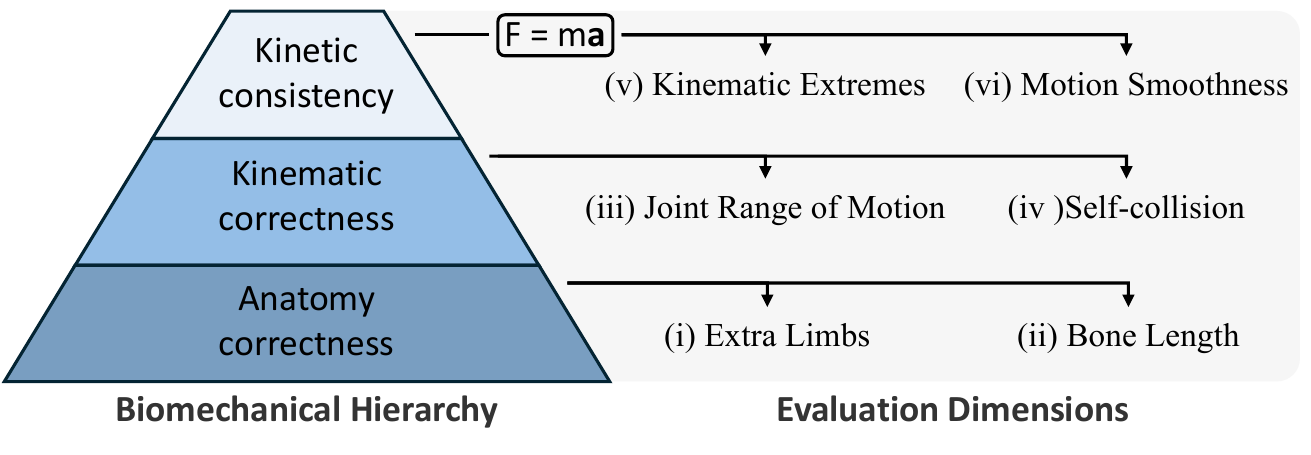}
    \caption{\textbf{Biomechanical hierarchy of our evaluation framework.} Each tier, from fundamental (bottom) to advanced (top), is evaluated using two independent metrics. \cradded{The top tier measures kinetic plausibility through temporal kinematics; no forces, torques, contacts, or inverse dynamics are estimated.}}
    \vspace{-1.0em}
    \label{fig:structure}
\end{wrapfigure}

\noindent\textbf{The big picture.} HumanScore follows a three-tier biomechanics hierarchy~\cite{hamill2006biomechanical}, as illustrated in \figureautorefname~\ref{fig:structure}. The base tier checks \emph{anatomical correctness} (valid topology and stable proportions), the middle tier checks \emph{kinematic correctness} (feasible pose geometry), and the top tier checks \emph{\cradded{kinetic plausibility}} (plausible temporal behavior). These tiers are dependent: anatomical failures can corrupt pose interpretation, and kinematic failures can make dynamic analysis unreliable. We therefore report six diagnostic dimensions rather than a single opaque score: spurious limbs or inconsistent proportions, hyperextension or self-collision, and non-human velocity, acceleration, or jerk artifacts.

\noindent\textbf{Reliance on external tools.} For AI-generated videos, motion-capture markers, IMUs, multi-view rigs, and force plates are unavailable, so monocular reconstruction is the only practical route to skeletons and meshes; several of our metrics therefore depend on these reconstructions and are inevitably exposed to estimation noise. We mitigate this in three ways. \emph{(i)} We adopt state-of-the-art monocular tools---HADM~\cite{Wang2024HADM}, MeTRAbs-XL~\cite{baumgartner2023monocular} with OpenSim~\cite{rajagopal2016full}, and GVHMR~\cite{shen2024gvhmr}---each near-saturating its own benchmark. \emph{(ii)} Reconstructions are treated as evidence rather than ground truth: every metric uses confidence-based filtering, biomechanics-derived tolerance margins, and the aggregation below, with the six dimensions reported separately so isolated fitting noise cannot dominate a composite score. \emph{(iii)} Each estimator is decoupled from the metric layer, so any future monocular estimator drops in and HumanScore tightens automatically. \sectionautorefname~\ref{sec:metric_robustness} verifies that rankings are stable under alternative pose pipelines and across tolerance and aggregation hyperparameters.
\cradded{Anatomical bounds come from biomechanics references; fitting tolerances are calibrated once on the Direct MoCap OpenSim reference set (100 trials, 17 subjects) to avoid penalizing valid measured motion, then frozen for all video evaluations. Collision filters suppress adjacent-face and small-area intersections.}
\emph{Per-metric implementation details are provided in the supplementary materials.}

\noindent\textbf{Robust video-level evaluations.} Five of our six metrics (the exception being bone length, which is a single time-aggregated deviation) share a common video-level protocol that summarizes per-frame violations along three orthogonal axes: \texttt{frequency} (how often a violation occurs), \texttt{severity} (how severe it is when it does), and \texttt{persistence} (how long it lasts). Concretely, each per-frame check yields a binary indicator $b_t\!\in\!\{0,1\}$ and a normalized severity $m_t\!\in\![0,1]$. Letting $T$ denote the number of frames and $L_{\max}$ the longest run of consecutive violations,
\[
\texttt{frequency}=\tfrac{1}{T}\textstyle\sum_t b_t,\quad
\texttt{severity}=\tfrac{\sum_t m_t}{\max(1,\,\sum_t b_t)},\quad
\texttt{persistence}=\tfrac{L_{\max}}{T},
\]
which combine into $\mathrm{Score}=100\,(1-\alpha\,\texttt{frequency}-\beta\,\texttt{severity}-\gamma\,\texttt{persistence})$ with $(\alpha,\beta,\gamma){=}(0.5,0.3,0.2)$. Higher scores indicate more biomechanically plausible motion. Sharing this aggregation keeps the six dimensions directly comparable. We analyze sensitivity to $(\alpha,\beta,\gamma)$ in supplementary materials.

\paragraph{\textbf{Anatomical Correctness.}} Anatomical correctness is the foundation of biomechanical evaluation, ensuring a stable topological structure across time. It guards against implausible phenomena such as duplicated arms/legs or frame-to-frame stretching and shrinking of bone segments, which would invalidate subsequent motion analysis. We check for \emph{extra limbs} and invariance of \emph{bone lengths}.

\noindent\texttt{(I) Extra limbs.} Anatomically impossible duplicate parts (e.g., an additional hand/arm/leg/foot or a ghost-like duplicate segment) violate human topology and invalidate any higher-level kinematic or kinetic assessment.
We adopt HADM~\cite{Wang2024HADM}, a detector trained on human images to flag structural anomalies, and combine its per-frame detection confidence with the relative area of the detected bounding boxes into a signal that we threshold and ramp into $b_t$ and $m_t$ for the aggregation above.

\noindent\texttt{(II) Bone length.} In real humans, bone-segment lengths are temporally constant (rigid-body constraint)~\cite{sun2017compositional}, so cross-frame variation in bone length is a strong indicator of anatomical inconsistency.
Building on Peiffer et al.~\cite{peiffer2025portable} and Metrabs~\cite{sarandi2020metrabs}, we construct a biomechanics-aware monocular skeleton fitting pipeline that fits per-frame OpenSim-compatible skeletons~\cite{opensim} with time-varying scale, allowing bone lengths to vary across frames. For each major bone, we compute the relative $\ell_1$ deviation from its temporal median; the per-bone average is then mapped to $[0,100]$ via a tolerance that absorbs residual fitting noise.

\paragraph{\textbf{Kinematic Correctness.}}
Kinematic correctness evaluates whether pose geometry is feasible given a valid anatomy. It constrains joint's DoFs to anatomical ranges to prevent hyperextension, folding, or other non-physiologic postures. It also rules out self-penetration between body parts—a failure mode that cannot be easily resolved through post-processing such as temporal smoothing. We design metrics for violations in \textit{joint range of motion} and \textit{self-collision} to evaluate kinematic correctness.

\noindent\texttt{(III) Joint range of motion.} We measure per-frame joint angles and check whether they remain within physiologically valid ranges; violations such as hyperextension or extreme flexion indicate geometrically infeasible postures even when anatomy is valid.
Reusing the OpenSim skeletons from above, we compare each joint's per-frame angle against anatomical bounds derived from biomechanics standards~\cite{rajagopal2016full} and relaxed by a tolerance that absorbs fitting uncertainty~\cite{catelli2019musculoskeletal}. The relative magnitude of each per-DoF violation defines $m_t$ and feeds the aggregation above.

\noindent\texttt{(IV) Self-collision.} Impossible interpenetrations between distinct body parts (e.g., forearm--torso, thigh--shank) are another signal of kinematic failure, complementing range-of-motion by catching geometric infeasibility that need not exceed any single joint limit.
We use GVHMR~\cite{shen2024gvhmr}, a monocular video fitter with strong temporal coherence, to recover per-frame SMPL-X meshes, and detect interpenetration via Bounding Volume Hierarchy (BVH) triangle-triangle tests with adjacency exclusion and non-local pair/area filters that suppress numerical noise and cloth contact. The fraction of colliding faces, ramped between mild and severe thresholds, defines $m_t$ for the aggregation, with severe-frame frequency driving the dominant term.

\paragraph{\textbf{\cradded{Kinetic Plausibility}.}}
Building on geometric feasibility, the top tier asks whether motion unfolds with plausible dynamics. \cradded{We measure higher-order kinematic consistency as a surrogate for dynamic plausibility, using velocity/acceleration-based diagnostics rather than force or torque estimates:} \textit{kinematic extremes} (joint angular and limb linear velocities) and \textit{motion smoothness} (acceleration regularity). Temporal derivatives are computed in physical time units using each video's native frame rate, so magnitudes remain directly comparable to per-DoF biomechanics limits.

\noindent\texttt{(V) Kinematic extremes.} A frame is suspicious if joint angular velocities or body-segment linear velocities exceed human capability limits, even when angles themselves remain within their range of motion.
From the fitted OpenSim skeletons, we obtain per-frame joint angular velocities via finite differences and body-segment linear velocities via forward kinematics at each segment's center of mass, then compare both against per-DoF and per-segment limits from biomechanics standards~\cite{sun2017compositional}. Normalized violation proportions, fused across joints and segments, define $m_t$ for the aggregation.

\noindent\texttt{(VI) Motion smoothness.} Natural human motion is smooth in time; excessive angular accelerations correspond to jitter, stutter, or discontinuities, signaling \cradded{temporal-plausibility failures}.
From the OpenSim angles, angular accelerations are obtained by cascaded finite differences. Following biomechanics standards~\cite{grimmer2020lowerlimb}, we compare per-frame accelerations against per-DoF limits and pass the normalized violation proportions through the same aggregation as the other metrics.

\noindent All six metrics report on a $[0,100]$ scale; \cradded{lower scores indicate greater measured biomechanical implausibility, subject to reconstruction uncertainty.} The six dimensions are reported separately rather than collapsed into one number, so any tier-specific failure mode remains diagnosable.

\begin{table*}[htb]
\caption{
\textbf{HumanScore Leaderboard.} Higher scores indicate better performance.
The best score in each dimension is highlighted by cell color. \cradded{Kinetic plausibility uses temporal kinematics, without force or torque estimation. Prompt-level SD/SEM and approximate intervals are in Appendix~\ref{sec:stats_app}; small score gaps should not be read as strict rankings.} The radar chart shows each model's overall performance across motion difficulty levels (Top) and over each metric (Bottom).
}

\renewcommand\tabcolsep{2pt}
\renewcommand\arraystretch{1.3}
\renewcommand{\thefootnote}{\fnsymbol{footnote}}
\setlength{\abovecaptionskip}{0.0cm}
\setlength{\belowcaptionskip}{-0.45cm}
\centering
\newsavebox{\leadertabbox}
\sbox{\leadertabbox}{\resizebox{0.7\linewidth}{!}{%
\begin{tabular}{r |
>{\centering\arraybackslash}p{1.2cm}
>{\centering\arraybackslash}p{1.2cm}
>{\centering\arraybackslash}p{1.2cm}|
>{\centering\arraybackslash}p{1.2cm}
>{\centering\arraybackslash}p{1.2cm}
>{\centering\arraybackslash}p{1.2cm}|
>{\centering\arraybackslash}p{1.2cm}
>{\centering\arraybackslash}p{1.2cm}
>{\centering\arraybackslash}p{1.2cm}|
>{\centering\arraybackslash}p{1.2cm}}
\toprule
\multicolumn{1}{c|}{\multirow{2}{*}{Models}} &
  \multicolumn{3}{c|}{\textbf{Anatomy Correctness}} &
  \multicolumn{3}{c|}{\textbf{Kinematic Correctness}} &
  \multicolumn{3}{c|}{\textbf{\cradded{Kinetic Plausibility}}} & \multirow{2}{*}{Overall}
   \\
\cmidrule(lr){2-4} \cmidrule(lr){5-7} \cmidrule(lr){8-10}
 & \texttt{(I)} & \texttt{(II)} & Avg & \texttt{(III)} & \texttt{(IV)} & Avg & \texttt{(V)} & \texttt{(VI)} & Avg &  \\
\hline
\hline
\rowcolor{row_color}
\multicolumn{1}{l|}{\emph{Proprietary models}} &&&&&&&&&&\\
\includegraphics[height=0.75em]{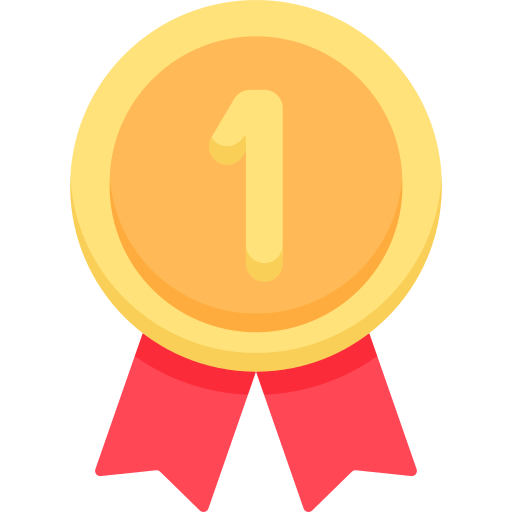} Seedance 1.0 Pro fast~\cite{gao2025seedance} & $94.2$ & $93.6$ & \thirdbest{$93.9$} & $86.8$ & $85.8$ & \secondbest{$86.3$} & $97.1$ & $98.1$ & \best{$97.6$} & \textbf{$92.6$} \\
\includegraphics[height=0.75em]{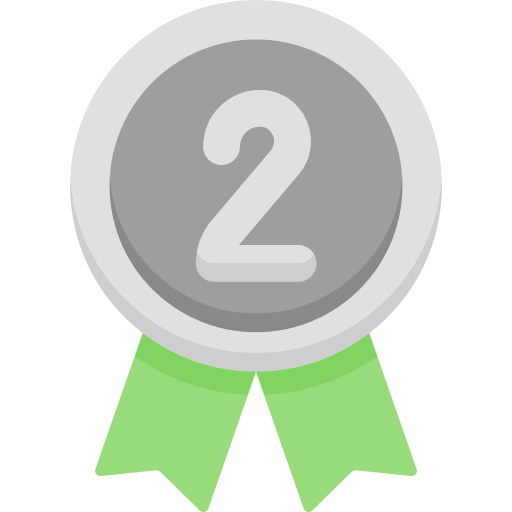} KlingAI 2.5 Turbo Pro~\cite{kling25turbo} & $89.3$ & $92.6$ & $91.0$ & $82.9$ & $90.3$ & \best{$86.6$} & $94.0$ & $95.9$ & $95.0$ & \textbf{$90.8$} \\
Ray 3.0~\cite{ray30} & $80.5$ & $92.8$ & $86.6$ & $78.6$ & $89.0$ & \thirdbest{$83.8$} & $95.2$ & $98.6$ & \thirdbest{$96.9$} & \textbf{$89.1$} \\
PixVerse 5.5~\cite{pixverse55} & $82.9$ & $91.0$ & $86.9$ & $72.4$ & $85.9$ & $79.1$ & $96.1$ & $98.3$ & \secondbest{$97.2$} & \textbf{$87.8$} \\
Hailuo 02~\cite{hailuo02} & $85.6$ & $92.5$ & $89.1$ & $77.5$ & $82.8$ & $80.2$ & $90.1$ & $96.5$ & $93.3$ & \textbf{$87.5$} \\
Veo 3.1 fast~\cite{googleveo31} & $78.4$ & $90.8$ & $84.6$ & $74.5$ & $87.5$ & $81.0$ & $94.7$ & $94.8$ & $94.7$ & \textbf{$86.8$} \\
Wan 2.6~\cite{wan26} & $85.8$ & $93.3$ & $89.6$ & $77.5$ & $87.9$ & $82.7$ & $84.9$ & $89.7$ & $87.3$ & \textbf{$86.5$} \\
Pika v2.2~\cite{pika22} & $86.0$ & $90.3$ & $88.2$ & $82.8$ & $82.5$ & $82.7$ & $78.3$ & $82.8$ & $80.5$ & \textbf{$83.8$} \\
Sora-2~\cite{openai_sora2_system_card_2025} & $91.9$ & $89.7$ & $90.8$ & $77.8$ & $83.8$ & $80.8$ & $73.9$ & $73.1$ & $73.5$ & \textbf{$81.7$} \\
\hline
\hline
\rowcolor{row_color}
\multicolumn{1}{l|}{\emph{Open-sourced models}} &&&&&&&&&&\\
\includegraphics[height=0.75em]{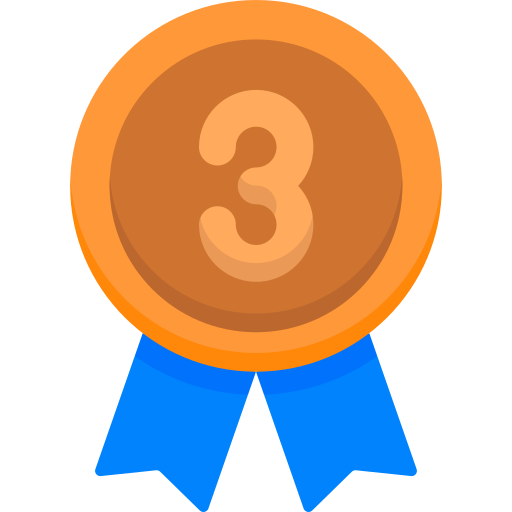}  HunyuanVideo 1.5~\cite{wu2025hunyuanvideo} & $95.6$ & $94.9$ & \best{$95.2$} & $80.6$ & $85.2$ & $82.9$ & $89.2$ & $90.1$ & $89.7$ & \textbf{$89.3$} \\
Wan 2.2~\cite{wan2025wan} & $96.1$ & $91.9$ & \secondbest{$94.0$} & $74.4$ & $85.7$ & $80.1$ & $81.2$ & $82.6$ & $81.9$ & \textbf{$85.3$} \\
Kandinsky 5.0 pro~\cite{kandinsky50} & $81.8$ & $91.6$ & $86.7$ & $71.8$ & $85.6$ & $78.7$ & $76.2$ & $78.9$ & $77.5$ & \textbf{$81.0$} \\
CogVideoX-5B~\cite{yang2024cogvideox} & $88.5$ & $59.1$ & $73.8$ & $58.9$ & $69.7$ & $64.3$ & $80.1$ & $92.5$ & $86.3$ & \textbf{$74.8$} \\
\hline
\hline
\textcolor{gray}{Real Videos} & \textcolor{gray}{$100$} & \textcolor{gray}{$92.0$} & \textcolor{gray}{$96.0$} & \textcolor{gray}{$99.9$} & \textcolor{gray}{$89.1$} & \textcolor{gray}{$94.5$} & \textcolor{gray}{$97.7$} & \textcolor{gray}{$98.6$} & \textcolor{gray}{$98.2$} & \textcolor{gray}{\textbf{$96.2$}} \\
\bottomrule
\end{tabular}%
}}
\noindent\begin{minipage}[c]{0.7\linewidth}
\centering
\usebox{\leadertabbox}
\end{minipage}%
\hfill
\begin{minipage}[c]{0.3\linewidth}
\centering
\includegraphics[width=\linewidth,height=\dimexpr\ht\leadertabbox+\dp\leadertabbox\relax,keepaspectratio]{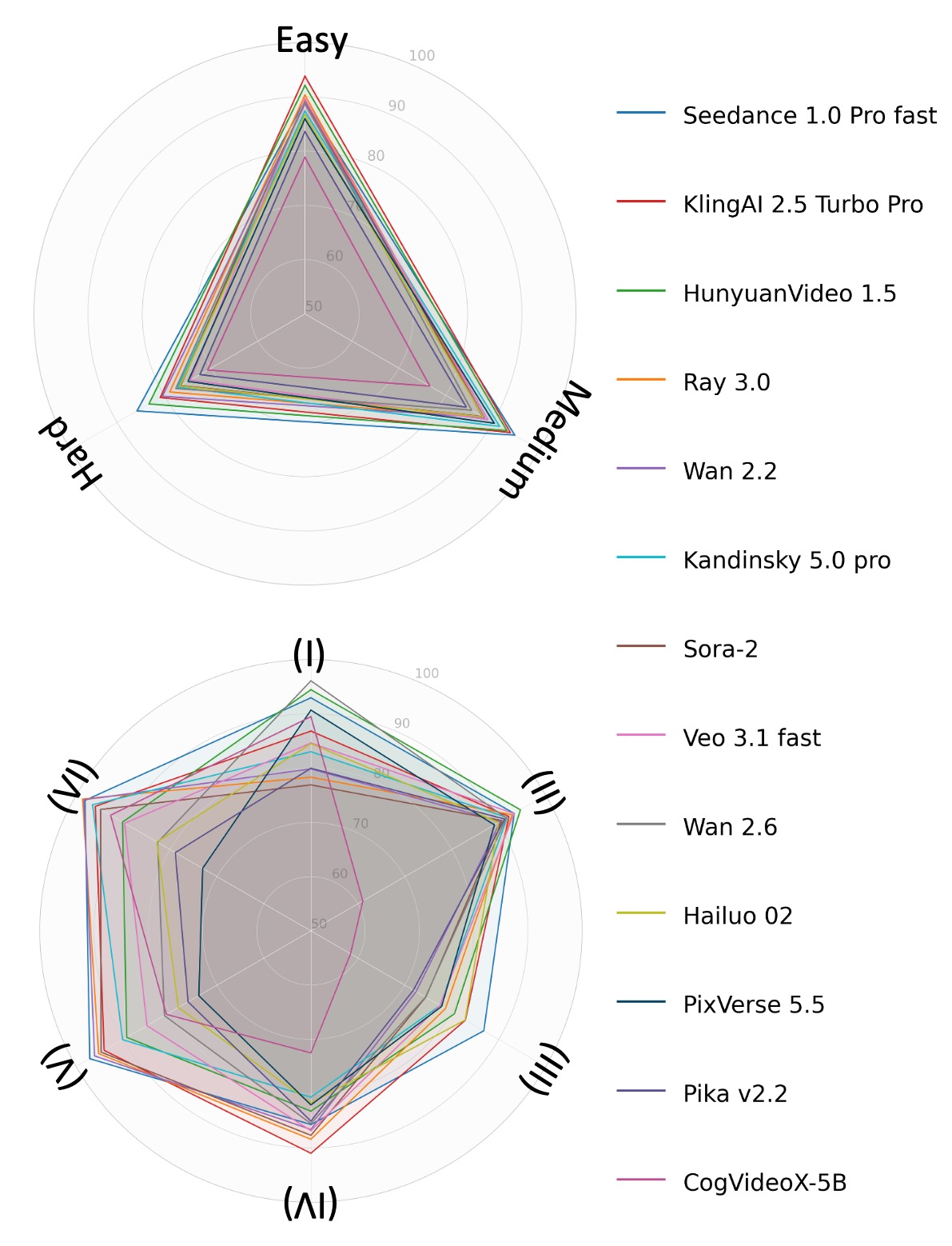}
\end{minipage}
\label{tab:model_compare}
\end{table*}

\section{Experiments and Results}
\label{sec:experiments}

\subsection{Verification of HumanScore Metrics on MoCap Data}


Because several HumanScore metrics rely on external monocular pose estimators, their measurements are exposed to estimation noise. To verify that the
  metrics are well-calibrated independent of reconstruction error, we first sanity-check them on directly measured MoCap motions. We use a Direct Mocap
  OpenSim reference set derived from the public AddBiomechanics dataset~\cite{werling2023addbiomechanics, werling2024addbiomechanicsdataset}. The source
  motions come from laboratory optical motion capture and are provided as OpenSim-compatible musculoskeletal trajectories, avoiding SMPL fitting,
  monocular reconstruction, or SMPL-to-OpenSim conversion. We select 100 trials from AddBiomechanics subjects and use the low-pass-filtered processing pass to obtain
  subject-specific OpenSim musculoskeletal models and time-varying generalized-coordinate trajectories.

  Across this Direct Mocap OpenSim set, we tune the skeleton-based metrics to produce a \textbf{near-perfect average score of 99.77\%} on directly
  measured human motion, and use the parameters unchanged for all subsequent evaluations. This reconstruction-free sanity check confirms that
  biomechanically valid motion is not penalized by overly strict thresholds, and provides the upper-bound anchor against which the in-the-wild and AI-
  generated results are interpreted. More details are provided in the supplementary materials.

\subsection{Evaluate Video Generation Models}
We benchmark thirteen video generation models, including four open-source systems—Wan 2.2~\cite{wan2025wan}, CogVideoX-5B~\cite{yang2024cogvideox}, HunyuanVideo 1.5~\cite{wu2025hunyuanvideo}, and Kandinsky 5.0 pro~\cite{kandinsky50}—and nine proprietary systems: Sora-2~\cite{openai_sora2_system_card_2025}, Veo 3.1 fast~\cite{googleveo31}, KlingAI 2.5 Turbo Pro~\cite{kling25turbo}, Seedance 1.0 Pro fast~\cite{gao2025seedance}, Hailuo 02~\cite{hailuo02}, Pika v2.2~\cite{pika22}, PixVerse 5.5~\cite{pixverse55}, Ray 3.0~\cite{ray30}, and Wan 2.6~\cite{wan26}. The models are evaluated using our proposed biomechanics-informed metrics across three major dimensions: anatomy correctness, kinematic correctness, and \cradded{kinetic plausibility}.

According to the leaderboard reported in \tableautorefname~\ref{tab:model_compare}, Seedance 1.0 Pro fast leads overall ($92.6$), \cradded{followed by KlingAI 2.5 Turbo Pro ($90.8$), HunyuanVideo 1.5 ($89.3$, strongest open-source model), and Ray 3.0 ($89.1$).} The per-dimension averages reveal complementary strengths. In \textbf{Anatomy Correctness}, HunyuanVideo 1.5 ($95.2$), Wan 2.2 ($94.0$), and Seedance ($93.9$) are strongest. In \textbf{Kinematic Correctness}, KlingAI ($86.6$) ranks first, followed by Seedance ($86.3$) and Ray 3.0 ($83.8$). In \textbf{\cradded{Kinetic Plausibility}}, Seedance ($97.6$), PixVerse 5.5 ($97.2$), and Ray 3.0 ($96.9$) achieve the best scores. Among the remaining models, PixVerse ($87.8$), Hailuo 02 ($87.5$), and Veo 3.1 fast ($86.8$) form a competitive middle tier, while Sora-2 ($81.7$), Kandinsky 5.0 pro ($81.0$), and especially CogVideoX-5B ($74.8$, with particularly weak anatomy and kinematic averages) trail the rest. The radar inset accompanying \tableautorefname~\ref{tab:model_compare} visualizes these per-tier scores as polygons, making the differing specialization patterns across generators easy to see at a glance. Performance also degrades systematically as motion difficulty increases (easy $\to$ medium $\to$ hard), supporting the role of the difficulty stratification (\sectionautorefname~\ref{sec:motion_curation}) in distinguishing biomechanical demand. \cradded{Examples include an inverted elbow in HunyuanVideo\textquotesingle s bending-back motion, an abrupt drop in Wan 2.6\textquotesingle s gymnastics tumbling, and duplicated limbs in Pika\textquotesingle s cartwheeling (Appendix~\ref{sec:qualitative_results}).}

\subsection{Correlation with Human Preference}

Beyond the quantitative leaderboard in \tableautorefname~\ref{tab:model_compare}, we conduct a human preference study to verify whether our biomechanics-informed metrics align with human judgment. We construct surveys with randomly paired videos generated from the same prompt by different models. In total, we collect $\mathbf{\sim1200}$ responses from researchers in both the AI and biomechanics communities. \cradded{Each response randomly assigns a video pair and an evaluation axis; participants watch both full videos and choose A, B, or a tie for that axis. Pairs are not systematically repeated across annotators, so inter-annotator agreement is not estimated.}

\begin{figure}[tb]
    \centering
    \includegraphics[width=0.99\linewidth]{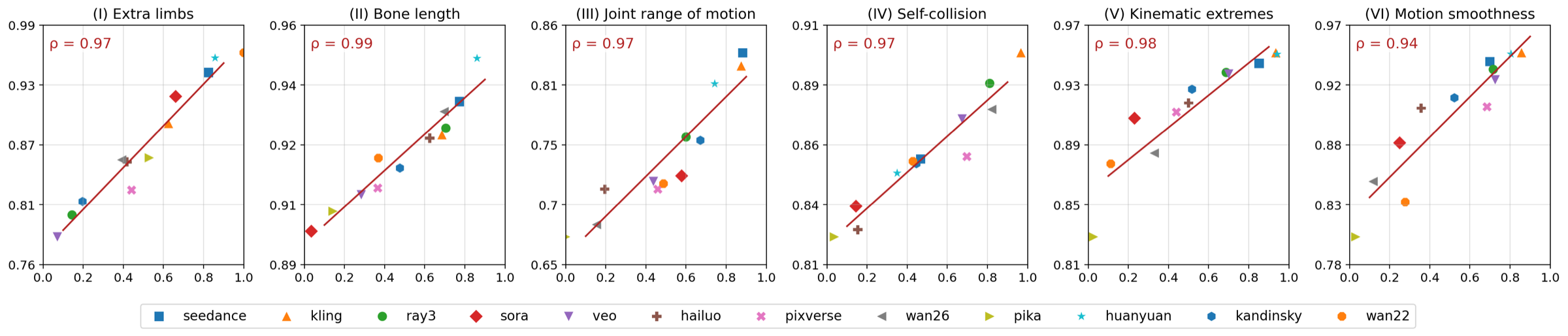}
    \caption{HumanScore highly aligns with human preference. The plots show averaged HumanScore win rate (Y) vs.\ human-preference win ratio (X), with a linear fit and Spearman's $\rho$.}
    \label{fig:preference}
    \vspace{-1em}
\end{figure}

Following VBench~\cite{10657096}, we score each pairwise comparison as 1 (win), 0.5 (tie), or 0 (loss), and compute each model's win ratio as its total score divided by the number of comparisons it appears in. We then compare model-level win ratios from human annotations with those from HumanScore, and report Spearman's rank correlation coefficient. As shown in \figureautorefname~\ref{fig:preference}, HumanScore exhibits strong agreement with human preference, with Spearman correlation coefficients close to 1.0 across evaluation dimensions. \cradded{These results support alignment with human judgment at the model-ranking level.}

\subsection{Real or AI?} \label{sec:realorai}

Here we revisit the challenge from \figureautorefname~\ref{fig:teaser}: can HumanScore distinguish real videos from AI-generated ones based on motion
  content alone? To anchor the metrics with physically measured human motion, we use a Direct Mocap OpenSim reference set derived from the public
  AddBiomechanics dataset~\cite{werling2023addbiomechanics, werling2024addbiomechanicsdataset}. Unlike SMPL-based or monocularly reconstructed motion,
  these sequences originate from laboratory optical motion capture and are already represented as OpenSim-compatible musculoskeletal trajectories.
  HumanScore averages \textbf{99.77} on this Direct Mocap OpenSim set, confirming that our metric thresholds do not reject valid human biomechanics.

  In-the-wild internet videos are a stricter test than direct MoCap captures, since their 3D motion must be recovered monocularly rather than measured
  directly. We collect such videos, each showing a single person performing a motion from our curated list (\figureautorefname~\ref{fig:prompt_stats}). \cradded{Selection matches action type and visibility/camera constraints, but not intensity (Appendix~\ref{sec:real_video_collection}).}
  Noteworthy, all metric thresholds are tuned once on Direct Mocap OpenSim and then fixed; HumanScore averages \textbf{96.2} on these real videos, higher
  than every benchmarked generator. Because no calibration is performed on this in-the-wild pool, the separation is not a fitting artifact: thresholds
  tuned on directly measured OpenSim motion generalize to noisy monocular reconstructions, supporting that despite the generators' strong visual realism,
  current systems still struggle with biomechanics-constrained motion.

  \cradded{The gap between these reference sets is consistent with reconstruction error, but is not a controlled estimate of that error because the motions differ:} internet videos are subject to depth
  ambiguity, occlusion, and motion blur~\cite{zheng2023poseSurvey,hmrKanazawa17,kocabas2019vibe}, and may include near-limit or atypical poses (e.g.,
  extreme flexibility) that legitimately fall outside conservative literature-derived bounds. These same factors explain why a few generators reach sub-
  metric values close to or marginally above real videos on individual dimensions (e.g., bone-length stability, self-collision); the overall ordering is
  nevertheless consistent; \cradded{$96.2$ is an empirical reference under this protocol, rather than a universal ceiling.}

\subsection{Correlation with Existing Benchmark Metrics} \label{sec:metric_correlation}
To clarify what HumanScore adds beyond existing benchmarks, we compute Spearman correlations between our three tier evaluation dimensions and three appearance-oriented axes from VBench~\cite{10657096} (imaging quality, aesthetic quality, and subject consistency). \tableautorefname~\ref{tab:metric_correlation} shows strong correlations for anatomy ($0.67$--$0.81$) and kinematics ($0.90$--$0.96$), but substantially weaker correlations for kinetics ($0.19$--$0.35$). The first two are expected: anatomical and pose-geometry failures tend to be visually conspicuous and are partially captured by frame-level visual quality measures. The kinetic disconnect, in contrast, shows that temporal dynamics are largely invisible to appearance-focused benchmarks. \cradded{HumanScore therefore complements the three VBench axes studied here, particularly in temporal motion plausibility.}

\begin{wraptable}{r}{0.39\textwidth}
    \vspace{-1.2em}
    \centering
    \small
    \setlength{\tabcolsep}{2pt}
    \resizebox{0.39\textwidth}{!}{%
    \begin{tabular}{lccc}
        \toprule
        & \begin{tabular}[c]{@{}c@{}}Imaging\\quality\end{tabular} & \begin{tabular}[c]{@{}c@{}}Aesthetic\\quality\end{tabular} & \begin{tabular}[c]{@{}c@{}}Subject\\consistency\end{tabular} \\
        \midrule
        Anatomy   & 0.669 & 0.679 & 0.808 \\
        Kinematic & 0.927 & 0.898 & 0.962 \\
        Kinetic   & 0.349 & 0.328 & 0.186 \\
        \bottomrule
    \end{tabular}%
    }
    \caption{Spearman correlations between our biomechanics-informed metrics and VBench~\cite{10657096} evaluation axes.}
    \label{tab:metric_correlation}
    \vspace{-1.2em}
\end{wraptable}

\subsection{Metric Robustness Analysis} \label{sec:metric_robustness}
We assess the stability of HumanScore along three axes: \textbf{(i)} \emph{choice of pose estimator}. Replacing our default pipeline with PromptHMR~\cite{wang2025prompthmr} and with a temporal-optimization variant on MeTRAbs keypoints; \textbf{(ii)} \emph{Scales of tolerances}. Tolerance parameters are set to absorb fitting noise, swept from strict to permissive; and \textbf{(iii)} \emph{aggregation hyperparameters} $(\alpha, \beta, \gamma)$ that balance frequency, severity, and persistence. They are swept across different combinations. \textbf{Quantitatively, in every condition tested the top-3 ordering of generators is preserved exactly.} No model moves by more than one position in the 13-model leaderboard, and Spearman rank correlation with our default ranking satisfies $\rho \ge 0.95$; the per-tier averages shift by less than $\pm 2$ points absolute on the $0$--$100$ scale, \cradded{which can exceed the gaps between closely ranked models in \tableautorefname~\ref{tab:model_compare}. These results support broad ranking stability under the tested choices, while reconstruction error remains a limitation.} \cradded{Appendix~\ref{sec:robustness_app} provides details, including reconstruction stress tests and repeated-generation results.}

\subsection{\cradded{Comparison with MLLM Evaluators}}\label{sec:mllm_comparison}
\cradded{GPT-5.6 Sol~\cite{openai2026gpt56sol} (image input) and Gemini 3.1 Pro Preview~\cite{google2026gemini31pro} (native video, without frame extraction) have model-level Spearman correlations of $0.715$ and $0.797$ with HumanScore. Gemini rates anatomical, kinematic, and kinetic plausibility on $0$--$100$ scales. Pearson correlations with HumanScore are $0.880$ and $0.782$, respectively. Gemini's narrower score range ($80.8$--$91.2$ versus $74.8$--$92.6$) suggests less separation between generators. Table~\ref{tab:mllm_scores} reports model means. The differing input modalities should be considered when comparing evaluators.}

\newpage
\noindent\begin{minipage}{\linewidth}
\centering
\begin{crrevision}
\captionsetup{hypcap=false}
\captionof{table}{\cradded{Model-level scores from HumanScore and MLLM evaluators. Higher scores indicate greater plausibility.}}
\label{tab:mllm_scores}
\small
\begin{tabular}{lrrr}
\toprule
Model & HumanScore & GPT-5.6 Sol~\cite{openai2026gpt56sol} & Gemini 3.1 Pro Preview~\cite{google2026gemini31pro} \\
\hline
\hline
\rowcolor{row_color}
\multicolumn{1}{l}{\emph{Proprietary models}} &&&\\
Seedance 1.0 Pro fast~\cite{gao2025seedance} & 92.6 & 85.3 & 90.8 \\
KlingAI 2.5 Turbo Pro~\cite{kling25turbo} & 90.8 & 82.6 & 91.2 \\
Ray 3.0~\cite{ray30} & 89.1 & 74.5 & 88.8 \\
PixVerse 5.5~\cite{pixverse55} & 87.8 & 76.3 & 85.4 \\
Hailuo 02~\cite{hailuo02} & 87.5 & 77.1 & 87.7 \\
Veo 3.1 fast~\cite{googleveo31} & 86.8 & 75.0 & 81.5 \\
Wan 2.6~\cite{wan26} & 86.5 & 75.9 & 84.2 \\
Pika v2.2~\cite{pika22} & 83.8 & 75.6 & 86.5 \\
Sora-2~\cite{openai_sora2_system_card_2025} & 81.7 & 70.3 & 81.7 \\
\hline
\hline
\rowcolor{row_color}
\multicolumn{1}{l}{\emph{Open-sourced models}} &&&\\
HunyuanVideo 1.5~\cite{wu2025hunyuanvideo} & 89.3 & 76.7 & 85.5 \\
Wan 2.2~\cite{wan2025wan} & 85.3 & 75.8 & 83.9 \\
Kandinsky 5.0 pro~\cite{kandinsky50} & 81.0 & 75.9 & 83.3 \\
CogVideoX-5B~\cite{yang2024cogvideox} & 74.8 & 62.8 & 80.8 \\
\bottomrule
\end{tabular}
\end{crrevision}
\end{minipage}
\par

\section{Discussions and Conclusions}
\label{sec:results}

\noindent\textbf{Findings.} Visual realism is saturating, but motion realism is not. Across 13 generators, all models still trail the real-video reference on HumanScore, despite the fact that top systems are already highly competitive on appearance-oriented benchmarks. In particular, our kinetic tier correlates only weakly with VBench's visual axes ($\rho{\le}0.35$, \tableautorefname~\ref{tab:metric_correlation}), suggesting that pixel-level or perceptual metrics provide limited signal for temporal-dynamic failures. Prompt engineering helps mitigate scene- and camera-level biases such as truncated bodies, moving cameras, and multi-person generations, but biomechanical violations remain visible even under controlled single-subject prompts. This indicates that motion plausibility cannot be fully solved at the input level and must instead be learned or enforced within the generative model.

Top generators also show distinct biomechanical profiles rather than uniform improvements. Seedance 1.0 Pro Fast~\cite{gao2025seedance} leads overall, \cradded{followed by KlingAI 2.5 Turbo Pro~\cite{kling25turbo} and HunyuanVideo 1.5~\cite{wu2025hunyuanvideo} (strongest open-source model), with Ray 3.0~\cite{ray30} close behind.} Their strengths differ across tiers: some models better preserve anatomical consistency, while others achieve stronger kinematic correctness or temporal stability. Meanwhile, highly dynamic generations often sacrifice bone-length consistency or smoothness, whereas conservative models tend to be smoother but under-actuated. This reveals a persistent trade-off between motion expressiveness and biomechanical stability, highlighting the need for explicit motion-level objectives in future video generators.


\noindent\textbf{Limitations.} \cradded{Reconstruction errors cannot be fully separated from generated-motion violations, especially with occlusion or visual artifacts. The controlled leaderboard covers single-person, full-body, static-camera videos; broader scene applicability remains preliminary.} With monocular inputs, forces, torques, and ground reactions cannot yet be reliably recovered, so \cradded{our kinetic plausibility tier uses higher-order kinematic surrogates rather than full kinetic measurements.} These catch the obvious temporal failures---jitter, spikes, drift---but leave subtler issues unaddressed, such as how ground reactions evolve during contact and how momentum transfers along kinematic chains. \cradded{We tested contact, surrogate inverse-dynamics, and support-polygon metrics, but excluded them because depth/contact uncertainty and legitimate aerial motion made them unreliable (Appendix~\ref{sec:additional_metrics}).}

\noindent\textbf{Conclusion.} We introduce HumanScore, a controlled biomechanics-grounded benchmark for human motion in AI-generated videos, covering 51 motion types and six interpretable metrics across anatomy, kinematics, and kinetics. Evaluating 13 state-of-the-art generators, HumanScore reveals motion-level gaps that visual benchmarks overlook: generated videos still lag behind real footage, kinetic realism remains weakly captured by appearance metrics, and leading models show different biomechanical strengths. We hope HumanScore can serve as a useful signal for building more physically plausible human-centric video models.

\newpage
\begin{ack}
\cradded{This work was partially funded by the NIH Grant R01AG089169 and P41EB027060, Panasonic Holdings Corporation, Stanford HAI, Stanford HAI graduate fellowship, Google cloud platform research credits, and Stanford Wu Tsai Human Performance Alliance.}
\end{ack}

\bibliographystyle{plainnat}
\begingroup
\let\croriginalbibitem\bibitem
\renewcommand{\bibitem}[2][]{%
  \color{black}\croriginalbibitem[#1]{#2}%
  \ifcamerareadyedits
    \ifcsname crbib@#2\endcsname\color{red}\fi
  \fi}
\bibliography{main}
\endgroup

\newpage
\appendix

\section{Teaser}

To facilitate comparison of the motions shown in \figureautorefname~\ref{fig:teaser}, we provide the corresponding videos in the supplementary material. We apply simple processing to better isolate the motion: specifically, we resize all videos to 480p and mask out the background. This keeps the person centered and reduces distracting factors that might bias judgment.

Alongside the processed videos, we include the corresponding metric scores. Our proposed metrics successfully detect biological and physical implausibility in the AI-generated videos; consequently, these videos receive lower scores than their real counterparts. This further supports the effectiveness of our metrics in distinguishing synthetic human motion from real motion.

\section{Availability and Broader Usage}

We evaluate all 13 generators on a fixed set of 102 prompts that span diverse motion types and difficulty levels. Sharing this prompt set across generators enables apples-to-apples comparison, but the benchmark is not the limit of HumanScore: the metrics it computes are motion-agnostic and apply to any monocular RGB video of a human.

\noindent\textbf{Extension to arbitrary motions.}
Our metrics rely on pose estimators and biomechanical constraints that are independent of motion semantics: no metric is calibrated to a specific action category, and no template motion is required as a reference. The evaluation pipeline therefore accepts arbitrary monocular RGB video as input without any curated prompt, and applies directly to in-the-wild videos covering motions well outside our 51-category set.

\noindent\textbf{Extension to multi-person settings.}
The benchmark prompt suite enforces a single subject per video to remove inter-person confounds, but the metric layer itself imposes no such restriction. Standard preprocessing---per-person detection, tracking, and segmentation---decomposes a multi-person clip into per-subject monocular streams, after which our evaluation runs unchanged on each subject.

Together, these two properties make HumanScore a controlled \emph{benchmark} built on top of a general-purpose \emph{metric suite}: the prompt suite gives a fair, apples-to-apples leaderboard, while the metric layer applies to any in-the-wild human video.

\section{Qualitative Fitting Results}\label{sec:qualitative_results}

Since the computation of our proposed metrics relies on fitted human models, we provide qualitative fitting results to help assess fit quality. For each model, we sample two videos with different motions: one with the highest HumanScore and one with the lowest HumanScore, and visualize the fitted skeletons throughout the sequence. We observe that our fitting pipeline captures the overall human motion well, even when the generated videos contain noticeable visual artifacts. This supports the robustness of the fitting pipeline and, in turn, the effectiveness of the metrics that depend on it. The best- and worst-HumanScore comparisons are provided in the \texttt{worst\_best} folder.

For poorly generated videos, the fitted skeleton often also appears distorted or implausible. This is expected: when the generated motion is unrealistic or the visual quality is low, even state-of-the-art fitters cannot recover a plausible human model, and our metrics accordingly assign lower scores. \cradded{Such cases require inspecting the video alongside the reconstruction, since a poor fit alone does not establish a generation failure.}

Note that, as shown in \autoref{fig:hmr_results} and discussed in \autoref{sec:future_work}, even state-of-the-art fitting methods, including those used in this work, remain fundamentally imperfect and cannot perfectly recover all human motions from monocular video. We expect future advances in human pose and shape estimation to further improve fitting quality and, as a result, make our metric computations even more reliable.

\section{Details on Metric Verification with MoCap Data} \label{sec:mocap_app}

We use a Direct Mocap OpenSim reference set derived from the public AddBiomechanics
dataset~\cite{werling2023addbiomechanics,werling2024addbiomechanicsdataset}. AddBiomechanics aggregates laboratory motion-capture datasets and provides processed
biomechanical representations, including subject-specific OpenSim-compatible models and joint kinematics. Unlike AIST++ or other SMPL-based references, this data
does not require mesh-to-skeleton conversion: the motion originates from experimental optical motion capture and is already available in a musculoskeletal
representation.

  Concretely, we select 100 trials from the \texttt{train/With\_Arm/Tiziana2019\_Formatted\_With\_Arm} subset of the AddBiomechanics data, covering 17 subjects.
Each trial is stored as a \texttt{.b3d} file. We load these files using NimblePhysics, select the AddBiomechanics low-pass-filtered processing pass, export the
corresponding subject OpenSim model, and write the generalized-coordinate trajectory as an OpenSim \texttt{.mot} file. The selected sequences have an average
sampling rate of approximately 61 Hz and are used directly as OpenSim trajectories for metric calibration and validation.

  Four of our metrics: \texttt{(II)} bone length, \texttt{(III)} joint range of motion, \texttt{(V)} kinematic extremes, and \texttt{(VI)} motion smoothness, consume per-frame OpenSim
skeleton trajectories with joint-specific degrees of freedom (DoFs) and biomechanically meaningful rotation conventions. The Direct Mocap OpenSim set is
therefore in the same representation consumed by these metrics, without passing through SMPL, monocular reconstruction, or SMPL-to-OpenSim conversion. This
avoids introducing artifacts from virtual marker placement, mesh fitting, or model retargeting, and provides a cleaner estimate of how valid human motion should
score under HumanScore.

  With these OpenSim trajectories, we tune the metric thresholds and aggregation settings that combine violation frequency, severity, and persistence. We
calibrate anatomical-limit tolerance, per-DoF velocity caps, and acceleration thresholds so directly measured human motion is not falsely flagged; selected
values stay within the ranges in \sectionautorefname~\ref{sec:metric_design_app} and are then fixed for all later experiments. After calibration, HumanScore
averages \textbf{99.77\%} on the Direct Mocap OpenSim reference set. The small residual loss comes from occasional near-limit poses, short high-velocity
transitions, or filtering artifacts in the processed mocap trajectories, which we consider acceptable because the metrics remain sensitive to genuine
biomechanical outliers.

\section{Details on Real-World Video Collection}\label{sec:real_video_collection}
As discussed in \sectionautorefname~\ref{sec:realorai}, to further validate our proposed metrics, we collect an extensive set of real-world videos from the internet. \cradded{These videos provide an end-to-end reference distribution that includes monocular reconstruction uncertainty.} Here we provide additional details and rationale for this collection process.

\cradded{We select one video for each of the 51 motion types in our curated list.} We begin by using these actions as search keywords on public video platforms (\cradded{YouTube (\url{https://www.youtube.com/})} and Bilibili~\cite{bilibili}) and in the YouTube-8M dataset~\cite{youtube8m}. We then retrieve candidate real-world videos and filter them using criteria consistent with those used to design the benchmark prompts. Specifically, we ensure that (1) the motion is clearly visible against a non-distracting background, (2) the full body remains visible throughout the video, (3) the camera viewpoint is stable so that subject motion is decoupled from camera motion, and (4) the video focuses on a single person. \cradded{Motion types are matched semantically; intensity is not matched to the gentle/intense prompts because it cannot be reliably inferred from real footage.}



\section{Details on the Fitting Process}

The metrics defined in \sectionautorefname~\ref{sec:metric_design} of the main paper draw on three external fitting pipelines, one per metric type: an \emph{anatomical-anomaly detector} for the topology check, a \emph{video-based SMPL-X mesh estimator} for self-penetration, and a \emph{biomechanics-aware skeleton fitter} for the four metrics that operate on per-frame OpenSim trajectories. We describe each below.

\subsection{Anatomical-Anomaly Detection \quad\small\emph{(Metric \texttt{(I)})}}
This metric detects whether humans in generated videos exhibit incorrect anatomies. We leverage HADM~\cite{Wang2024HADM}, a state-of-the-art model trained on both normal and anomalous human images. Its authors curated an in-house dataset of tens of thousands of images from major text-to-image systems, with detailed annotations for local artifacts (e.g., in faces, arms, hands, legs, feet) and global artifacts (e.g., missing or extra limbs). HADM is built on a ViTDet architecture, with false positives further reduced by training on real-human datasets that contain no artifacts, teaching the model what ``normal'' humans look like. HADM achieves state-of-the-art performance against prior artifact detectors and large vision-language models when ranking images by artifact severity, and generalizes well to images from unseen generators and to real photos.

\subsection{SMPL-X Mesh Recovery \quad\small\emph{(Metric \texttt{(IV)})}}
To detect kinematic failures such as self-penetration, generated motion must be analyzed at the body-surface level in 3D. We use GVHMR~\cite{shen2024gvhmr}, a video-based monocular HMR method that recovers per-frame SMPL-X meshes with strong temporal coherence. By jointly modeling each clip's global trajectory together with per-frame pose, GVHMR substantially reduces the foot-sliding and per-frame jitter that are common to single-image HMR methods---both of which would otherwise inflate spurious self-intersection counts. The temporally coherent meshes then feed directly into the BVH-accelerated triangle--triangle intersection tests described in \sectionautorefname~\ref{sec:metric_design} of the main paper.

\subsection{Biomechanics-Aware Skeleton Fitting \quad\small\emph{(Metrics \texttt{(II)}, \texttt{(III)}, \texttt{(V)}, \texttt{(VI)})}}
\begin{figure*}[t]
    \centering
    \includegraphics[width=0.99\linewidth]{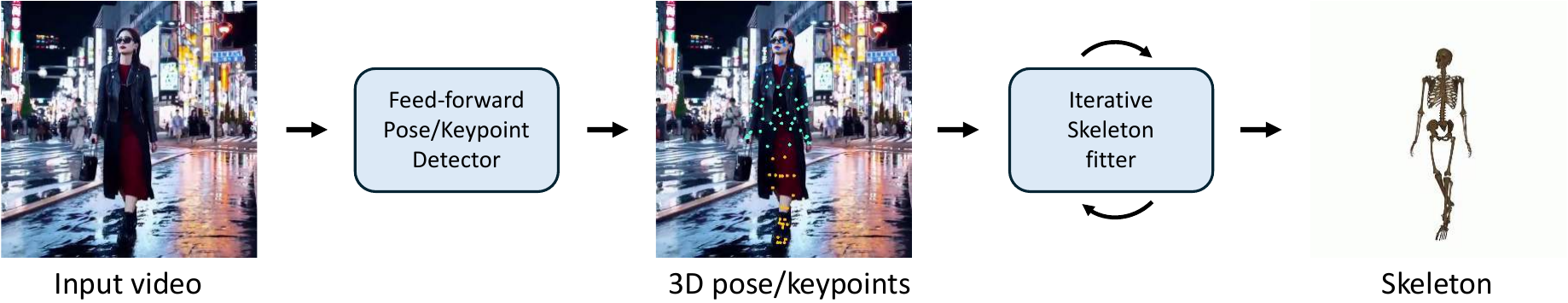}
    \caption{For metrics that rely on skeletal fitting, we use a two-stage pipeline. First, we apply a pretrained pose/keypoint detector to infer 87 3D keypoints from monocular observations. Second, we fit a biomechanics-informed human skeleton model to these keypoints via iterative optimization.}
    \label{fig:fitting_pipeline}
\end{figure*}

The remaining four metrics---bone length \texttt{(II)}, joint range of motion \texttt{(III)}, kinematic extremes \texttt{(V)}, and motion smoothness \texttt{(VI)}---rely on per-frame OpenSim trajectories with joint-specific degrees of freedom and rotation orders, a representation that end-to-end pose estimators do not provide directly. We therefore adopt a two-stage skeleton-fitting pipeline, illustrated in \figureautorefname~\ref{fig:fitting_pipeline}.

\paragraph{Stage 1: Feed-forward Keypoint Detector.} We use the pretrained MeTRAbs-XL model~\cite{sarandi2020metrabs,baumgartner2023monocular} to infer 87 anatomical 3D landmarks per frame. The commonly used 17~\cite{lin2014microsoft} or 24~\cite{loper2023smpl} keypoint conventions place a single keypoint per joint, which is sufficient for many perception tasks but does not constrain joint rotations or local bone orientations. With three or more landmarks per body segment, MeTRAbs-XL recovers all DoFs of a human skeletal model and provides the dense observations needed by Stage 2.

\paragraph{Stage 2: Iterative Skeleton Fitter.} We initialize a physics-based skeleton in a T-pose, read out the corresponding bone locations from the skeleton, and iteratively minimize the L2 distance between matched skeleton anchors and the per-frame keypoints from Stage 1~\cite{peiffer2025portable}. Optimization runs for 1000 steps per frame ($\approx$5\,s per frame on a single GPU).

\paragraph{Postprocessing: Alignment to OpenSim.} The fitted model's generalized coordinates are mapped to OpenSim joints, applying knee-sign corrections and a fixed axis reordering with Euler-angle conversion for pelvis translation and rotation. The resulting time series of joint angles and pelvis pose is exported as an OpenSim-compatible trajectory, directly consumable by metrics \texttt{(II)}, \texttt{(III)}, \texttt{(V)}, and \texttt{(VI)}.

\subsection{OpenSim Skeleton}
The OpenSim Rajagopal Model~\cite{rajagopal2016full} used in this work is an open-source biomechanical model, which represents the human skeleton as a set of rigid body segments connected by joints. The skeleton consists of the following rigid body segments:
\begin{itemize}
    \item Pelvis
    \item Right and left femur
    \item Right and left patella
    \item Right and left tibia/fibula
    \item Right and left talus
    \item Right and left calcaneus (including tarsal and metatarsal bones)
    \item Right and left toes
    \item Head and torso
    \item Right and left humerus
    \item Right and left ulna
    \item Right and left radius
    \item Right and left hand
\end{itemize}

\noindent These body segments are articulated by ball-and-socket, pin, and universal joints with anatomically accurate degrees of freedom (DoFs). Unlike SMPL~\cite{SMPL:2015, SMPL-X:2019}, where all joints uniformly have 3 DoFs, the OpenSim model assigns joint-specific DoFs that better represent the actual biomechanical constraints of the human body and ultimately facilitate the computation of our metrics. The skeleton consists of the following joints:
\begin{itemize}
    \item {Pelvis (root of the skeleton):} 6 DoFs --- translation and rotation with respect to the ground.

    \item {Hip:} ball-and-socket joint with 3 DoFs per leg --- flexion/extension, adduction/abduction, and internal/external rotation.

    \item {Knee:} pin joint with 1 DoF per leg --- flexion/extension.

    \item {Ankle:} pin joint with 1 DoF per leg --- dorsiflexion/plantarflexion.

    \item {Subtalar:} pin joint with 1 DoF per leg --- inversion/eversion.

    \item {Metatarsophalangeal:} pin joint with 1 DoF per leg --- toe flexion/extension.

    \item {Lumbar:} ball-and-socket joint with 3 DoFs --- extension/flexion, bending, and rotation.

    \item {Shoulder:} ball-and-socket joint with 3 DoFs per arm --- flexion/extension, adduction/abduction, and internal/external rotation.

    \item {Elbow:} pin joint with 1 DoF per arm --- flexion/extension.

    \item {Radioulnar:} pin joint with 1 DoF per arm --- pronation/supination.

    \item {Wrist:} universal joint with 2 DoFs per arm --- flexion/extension and ulnar/radial deviation.
\end{itemize}

\section{Details on Metric Design} \label{sec:metric_design_app}

\subsection{Metric Calculation} \label{sec:supp_metric}

In \sectionautorefname~\ref{sec:metric_design}, we present the six metrics used in our benchmark to score each video and assess whether the human motion is real or AI-generated. Here, we provide more details on how to compute these metrics.

\paragraph{Generic framework.}
To measure per-frame human motion behavior across a temporal sequence, it is important to have a robust and flexible framework for metric calculation. To this end, we follow a generic framework for computing all six metrics proposed in this work. For all the metrics, higher scores indicate more realistic motion.

These six metrics focus on different tiers in the biomechanical hierarchy (\figureautorefname~\ref{fig:structure}), and therefore each metric has its own \textbf{frame-level calculation}. The per-frame calculation is designed to produce three types of outputs: $b_t$ -- a binary indicator that flags whether a frame is normal or abnormal; $m_t$ -- a relative metric score that measures how ``bad'' a frame is, in the range $(0, 1)$; and $L_{\max}$ -- the length of the longest consecutive abnormal (according to $b_t$) frame segment.

For \textbf{video-level aggregation}, we seek to obtain a single numeric value as a unified score that represents the entire video from the three aspects of \texttt{frequency}, \texttt{severity}, and \texttt{persistence}, which can be computed and combined from the three outputs as:
\[
\texttt{frequency}=\tfrac{1}{T}\textstyle\sum_t b_t,\quad
\texttt{severity}=\tfrac{\sum_t m_t}{\max(1,\,\sum_t b_t)},\quad
\texttt{persistence}=\tfrac{L_{\max}}{T},
\]
where $T$ is the total number of frames. We define these three core scores as a more comprehensive summarization of the per-frame results. By balancing the $(\texttt{frequency},\texttt{severity},\texttt{persistence})$ triplet with the corresponding weights $\alpha{=}0.5,\ \beta{=}0.3,\ \gamma{=}0.2$ as hyper-parameters (see \autoref{sec:metric_robustness} for study on hyper-parameter robustness), we can obtain the final score for each metric.

\paragraph{\texttt{(I) Extra limbs.}} 
We first run HADM~\cite{Wang2024HADM} on each frame, which produces candidate boxes for four limb classes: \{extra hand, arm, leg, foot\} with associated confidences.

\noindent\emph{Frame-level calculation.} Let $p_{\max}$ be the maximum confidence over the four classes at frame $t$, and $a$ the relative bounding-box area covered by extra-limb detections. We gate weak signals with $\tau_{\text{prob}}{=}0.005$ and $\tau_{\text{area}}{=}0.01$ and define $M_t = p_{\max} \cdot a$ if $p_{\max}\ge\tau_{\text{prob}}$ and $a\ge\tau_{\text{area}}$, otherwise $M_t = 0$. With $\tau_{\text{mild}}{=}0.005$ and $\tau_{\text{severe}}{=}0.03$, the per-frame outputs are:
\[
b_t =
\begin{cases}
0, & M_t \le \tau_{\text{mild}},\\
1, & \text{otherwise},
\end{cases}
\]
\[
m_t =
\begin{cases}
0, & M_t \le \tau_{\text{mild}},\\
\dfrac{M_t - \tau_{\text{mild}}}{\tau_{\text{severe}} - \tau_{\text{mild}}}, & \tau_{\text{mild}} < M_t < \tau_{\text{severe}},\\
1, & M_t \ge \tau_{\text{severe}}.
\end{cases}
\]
Here, $b_t$ is a binary indicator that shows whether the frame has a strong enough detection of anomalous limbs, and $m_t$ is the per-frame severity score in the range $[0,1]$.

\noindent\emph{Video-level aggregation.} We then calculate the aggregated scores over all frames:
\[
\texttt{frequency} = \tfrac{1}{T}\textstyle\sum_{t=1}^{T} b_t,\quad
\texttt{severity} = \tfrac{\sum_{t=1}^{T} m_t}{\max\!\bigl(1,\ \sum_{t=1}^{T} b_t\bigr)},\quad
\texttt{persistence} = \tfrac{L_{\max}}{T},
\]
where $T$ is the total number of frames and $L_{\max}$ is the length of the longest consecutive abnormal segment (defined using $b_t$) to better capture motion-blur cases.

Finally, the video-level metric is given by a weighted sum of the three aggregated scores:
\[
\mathrm{Score} = 100\,\bigl(1-\alpha\,\texttt{frequency}-\beta\,\texttt{severity}-\gamma\,\texttt{persistence}\bigr),
\]
where $\alpha{=}0.5$, $\beta{=}0.3$, and $\gamma{=}0.2$ are hyper-parameters. Higher scores indicate more anatomically natural human bodies.

\paragraph{\texttt{(II) Bone length.}} 
This metric requires fitted OpenSim skeletons for all frames in the video.

\noindent\emph{Frame-level calculation.} At frame $t$, for a bone $b=(\text{parent},\,\text{child})$ sampled from the fitted skeleton, the bone length is defined as:
\[
l_b(t)=\bigl\|\mathbf{p}_{\text{parent}}(t)-\mathbf{p}_{\text{child}}(t)\bigr\|_2.
\]
For this metric, we discard frames with low keypoint detection confidence or large skeleton reprojection error, which indicate inaccurate bone length estimation.

Given the unstable nature of AI-generated skeleton and thus the bone length, for each bone we use the median length across the entire video as a robust reference:
\[
L_b=\operatorname{median}_t\, l_b(t),\qquad
e_b(t)=\frac{\bigl|\,l_b(t)-L_b\,\bigr|}{L_b+\epsilon},
\]
where $\epsilon{=}10^{-8}$ is a small constant to avoid division by zero. For each bone $b$, we then define the bone-level average error $E_b$ across all frames:
\[
E_b=\frac{1}{T}\sum_{t=1}^{T} e_b(t).
\] 

\noindent\emph{Video-level aggregation.} To aggregate bone-level errors, we compute the average error ${\bar E}$ over the nine major bones (femur/tibia/humerus/forearm L\&R and torso), and then map this average to a $[0,100]$ scale using a tolerance hyper-parameter $\tau{=}0.40$:
\[
\mathrm{Score} = 100\left(1-\operatorname{clip}\!\left(\frac{\bar E}{\tau},\,0,\,1\right)\right).
\]
Higher scores indicate more stable bone lengths and stronger anatomical consistency. For robustness, we require at least ten valid frames per segment, and use the per-bone median (not mean) as the temporal reference, which is robust to per-frame fitting outliers; torso-scale cross-checks (e.g., pelvis--neck) help disentangle genuine depth changes from local stretch artifacts.

\paragraph{\texttt{(III) Joint range of motion.}}
For each joint $j$ and DoF $d$ read from the fitted OpenSim skeleton, we obtain a sequence $\theta_{t}^{(j,d)}$ over frames $t$. This metric checks whether any time step of $\theta_{t}^{(j,d)}$ violates the predefined joint anatomical limits $[\theta_{\min}^{(j,d)},\,\theta_{\max}^{(j,d)}]$ based on biomechanics standards~\cite{rajagopal2016full}. These limits are relaxed by a tolerance $\mathrm{tol}=15^{\circ}$, following~\cite{catelli2019musculoskeletal}, to account for fitting uncertainty.

\noindent\emph{Frame-level calculation.}
The degree of violation for $\theta_{t}^{(j,d)}$ is defined as:
\[
\Delta_{t}^{(j,d)}=\max\!\Bigl\{0,\ \theta_{t}^{(j,d)}-(\theta_{\max}^{(j,d)}+\mathrm{tol}),\ (\theta_{\min}^{(j,d)}-\mathrm{tol})-\theta_{t}^{(j,d)}\Bigr\}.
\]
Let $\delta_{j,d}=0.5\!\cdot\!(\theta_{\max}^{(j,d)}-\theta_{\min}^{(j,d)})$ be the relative range of the limit. We obtain the relative extent of violation by scaling:
\[
\begin{aligned}
m_t&=\min\!\bigl(\Delta_{t}^{(j,d)}/\delta_{j,d},\,1\bigr)\ \in[0,1],\\
b_t&=\mathbbm{1}[\,m_t>\tau\,],
\end{aligned}
\]
where $b_t$ is a per-frame, per-joint indicator for range-of-motion violation and $\tau{=}0.05$ is a violation-threshold hyper-parameter.

\noindent\emph{Video-level aggregation.}
Similar to other metrics, we calculate the averages of $b_t$ and $m_t$ over all frames:
\[
\texttt{frequency}=\tfrac{1}{T}\textstyle\sum_{t=1}^{T} b_t,\quad
\texttt{severity}=\tfrac{\sum_{t=1}^{T} m_t}{\max\!\bigl(1,\sum_{t=1}^{T} b_t\bigr)},\quad
\texttt{persistence}=\tfrac{L_{\max}}{T},
\]
where $L_{\max}$ is the length of the longest consecutive violated segment, which helps capture temporally persistent artifacts such as motion blur.

Finally, the score for this metric is given by the weighted sum of the three aggregated quantities:
\[
\mathrm{Score}=100\,\bigl(1-\alpha\,\texttt{frequency}-\beta\,\texttt{severity}-\gamma\,\texttt{persistence}\bigr),
\]
where $\alpha{=}0.5$, $\beta{=}0.3$, and $\gamma{=}0.2$ are hyper-parameters that are kept consistent across all metrics. Higher scores indicate motions that are more consistent with normal joint ranges of motion.

\paragraph{\texttt{(IV) Self collision.}}
This metric requires accurate SMPL-X meshes for all frames in the video. We use GVHMR~\cite{shen2024gvhmr}, a state-of-the-art monocular video fitter that leverages global trajectories and temporal coherence, to obtain reliable per-frame SMPL-X meshes.

\noindent\emph{Frame-level calculation.}
Let $F$ be the total number of mesh faces and $F_{\text{collide}}(t)$ the set of colliding faces at frame $t$. We define
\[
M_t=\frac{|F_{\text{collide}}(t)|}{F}\in[0,1].
\]
We then apply a double-threshold ramp with hyper-parameters $\tau_{\text{mild}}{=}0.01$ and $\tau_{\text{severe}}{=}0.03$:
\[
m_t=\min\!\left\{1,\ \max\!\left\{0,\ \frac{M_t-\tau_{\text{mild}}}{\tau_{\text{severe}}-\tau_{\text{mild}}}\right\}\right\},
\]
\[
b_t=\mathbbm{1}[\,m_t>0\,].
\]
Here, $m_t\in[0,1]$ is the per-frame collision severity, and $b_t$ is a binary indicator that marks frames with non-zero collision.

\noindent\emph{Video-level aggregation.}
Let $T$ be the total number of frames, and let $L_{\max}$ be the length of the longest consecutive segment with $b_t{=}1$ (i.e., frames with collisions), which captures persistent artifacts. We first define
\[
\texttt{severity}=\tfrac{\sum_{t=1}^{T} m_t}{\max\!\bigl(1,\sum_{t=1}^{T} \mathbbm{1}[\,m_t>0\,]\bigr)},\quad
\texttt{persistence}=\tfrac{L_{\max}}{T}.
\]
We also measure the fraction of frames with severe collisions:
\[
\texttt{frequency}_{\text{severe}}=\tfrac{1}{T}\textstyle\sum_{t=1}^{T}\mathbbm{1}\bigl[\,M_t\ge\tau_{\text{severe}}\,\bigr].
\]

We also track the mild-rate axis $\texttt{frequency}_{\text{mild}}{=}(1/T)\sum_t \mathbbm{1}[\,m_t{>}0\,]$ as an optional fourth axis. The final metric value is obtained from a weighted combination of these aspects:
\[
\mathrm{Score}=100\cdot\bigl(1-\operatorname{clip}\bigl(\alpha\,\texttt{frequency}_{\text{severe}}+\beta\,\texttt{severity}+\gamma\,\texttt{persistence}+\delta\,\texttt{frequency}_{\text{mild}},\,0,\,1\bigr)\bigr).
\]
The weights $(\alpha,\beta,\gamma,\delta){=}(0.5,0.3,0.2,0.0)$ emphasize severe and persistent collisions; the mild-rate axis is reported but not used by default ($\delta{=}0$). Higher scores indicate fewer and milder self-collisions.

\paragraph{\texttt{(V) Kinematic extremes.}} 
From the fitted per-frame OpenSim skeletons, similar to the other metrics, we obtain joint angle sequences $\theta^{(j,d)}_t$. We then compute joint angular velocities $\omega^{(j,d)}_t$ via first-order finite differences (with midpoint timestamps and gap filtering), and body-segment linear velocities $v^{(s)}_t$ via forward kinematics at each segment's center of mass. Both $\omega^{(j,d)}_t$ and $v^{(s)}_t$ are examined against per-DoF and per-segment limits $\omega^{\max}_{j,d}$ and $v^{\max}_s$ derived from biomechanics standards~\cite{sun2017compositional} \cradded{with fitting tolerances calibrated on the Direct MoCap OpenSim reference set} (\sectionautorefname~\ref{sec:experiments}).

\noindent\emph{Frame-level calculation.}
Similar to the joint range-of-motion violation metric, we first define the relative violation ratios per joint and per body segment, clamped with a fixed normalization factor of $0.5$:
\[
\begin{aligned}
m^{\text{joint}}_t&=\frac{\sum_{j,d}\,
\min\!\Bigl\{1,\ \max\!\bigl(0,\tfrac{|\omega^{(j,d)}_t|}{\omega^{\max}_{j,d}}-1\bigr)/0.5\Bigr\}}
{\sum_{j,d} w_{j,d}},
\end{aligned}
\]
\[
\begin{aligned}
m^{\text{body}}_t&=\frac{\sum_{s}\,
\min\!\Bigl\{1,\ \max\!\bigl(0,\tfrac{v^{(s)}_t}{v^{\max}_{s}}-1\bigr)/0.5\Bigr\}}
{\sum_{s} u_{s}},
\end{aligned}
\]
where $w_{j,d}$ and $u_s$ are anatomy-tiered weights: hip/knee/ankle/subtalar/MTP joints and pelvis/torso/head segments receive higher weights ($\sim$1.5--2.0), trunk and limb segments intermediate weights, and remaining DOFs default to 1.0.

The joint and body violation ratios are then fused at the frame level and counted as a concrete violation using a threshold $\tau{=}0.05$. If both signals are non-zero on a given frame they are averaged; otherwise the single non-zero term is used directly:
\[
\begin{aligned}
m_t&=\begin{cases}\tfrac{1}{2}(m^{\text{joint}}_t+m^{\text{body}}_t),&m^{\text{joint}}_t,m^{\text{body}}_t>0\\m^{\text{joint}}_t+m^{\text{body}}_t,&\text{otherwise}\end{cases}\\
b_t&=\mathbbm{1}[\,m_t>\tau\,].
\end{aligned}
\]

\noindent\emph{Video-level aggregation.}
We first compute the unified $(\texttt{frequency},\texttt{severity},\texttt{persistence})$ triplet:
\[
\texttt{frequency}=\tfrac{1}{T}\textstyle\sum_{t=1}^{T} b_t,\quad
\texttt{severity}=\tfrac{\sum_{t=1}^{T} m_t}{\max\!\bigl(1,\sum_{t=1}^{T} b_t\bigr)},\quad
\texttt{persistence}=\tfrac{L_{\max}}{T},
\]
where $T$ is the total number of frames and $L_{\max}$ is the length of the longest consecutive violated segment, capturing persistent extreme motions.

The final score is again a weighted sum of $(\texttt{frequency},\texttt{severity},\texttt{persistence})$ and is linearly mapped to the range $[0,100]$, where higher values indicate more natural motion:
\[
\mathrm{Score}=100\,\bigl(1-\alpha\,\texttt{frequency}-\beta\,\texttt{severity}-\gamma\,\texttt{persistence}\bigr).
\]

\paragraph{\texttt{(VI) Motion smoothness.}}
Starting from the fitted OpenSim skeletons, similar to the calculation of joint velocities above, we compute angular accelerations $\alpha^{(j,d)}_t$ via cascaded first-order finite differences (forward differences applied twice, with gap filtering against missing frames). We compare the magnitudes $|\alpha^{(j,d)}_t|$ to per-DoF limits $\alpha^{\max}_{j,d}$ derived from biomechanics standards~\cite{grimmer2020lowerlimb} \cradded{with fitting tolerances calibrated on the Direct MoCap OpenSim reference set}.

\noindent\emph{Frame-level calculation.}
We first compute the per-frame relative violation ratio for acceleration:
\[
q^{(j,d)}_t=\min\!\Bigl\{1,\ \max\!\bigl(0,\tfrac{|\alpha^{(j,d)}_t|}{\alpha^{\max}_{j,d}}-1\bigr)/0.5\Bigr\}.
\]
We then combine these into a single per-frame score using anatomy-tiered weights $w_{j,d}$ (hip/knee/ankle/MTP$\,{=}\,2.0$, lumbar/pelvis$\,{=}\,1.5$, arm/elbow/wrist$\,{=}\,1.2$, default$\,{=}\,1.0$) and a threshold $\tau{=}0.05$:
\[
m_t=\frac{\sum_{j,d} w_{j,d}\,q^{(j,d)}_t}{\sum_{j,d} w_{j,d}}\in[0,1],
\]
\[
b_t=\mathbbm{1}[\,m_t>\tau\,].
\]

\noindent\emph{Video-level aggregation.}
The video-level score is then computed across the entire sequence. Let $T$ be the total number of frames and $L_{\max}$ be the length of the longest consecutive segment with $b_t{=}1$ (i.e., frames with excessive acceleration):
\[
\texttt{frequency}=\tfrac{1}{T}\textstyle\sum_{t=1}^{T} b_t,\quad
\texttt{severity}=\tfrac{\sum_{t=1}^{T} m_t}{\max\!\bigl(1,\sum_{t=1}^{T} b_t\bigr)},\quad
\texttt{persistence}=\tfrac{L_{\max}}{T}.
\]

Finally, we obtain the overall metric:
\[
\mathrm{Score}=100\cdot\bigl(1-\operatorname{clip}\bigl(\alpha\,\texttt{frequency}+\beta\,\texttt{severity}+\gamma\,\texttt{persistence},\,0,\,1\bigr)\bigr),
\]
where $\alpha$, $\beta$, and $\gamma$ are shared hyper-parameters as in the other metrics. Higher scores indicate smoother and more physically plausible motion.

\subsection{Alternative Metric Implementations}
The implementation routes presented in \sectionautorefname~\ref{sec:metric_design} and \sectionautorefname~\ref{sec:supp_metric} are those we found to be the most reasonable and effective. During preliminary experiments, we exhaustively tested multiple alternative implementation routes for each metric. Here, we discuss these alternative implementations.

\paragraph{\texttt{(I) Extra limbs.}}
\begin{itemize}\itemsep 2pt
\item \textbf{DensePose-style instance IUV.}
One option is to detect extra limbs by counting connected components per limb inside a single ROI. However, single-instance matching often \emph{selects or averages one hypothesis}, which can hide duplicated limbs under motion blur. This approach also requires heavy per-pixel inference and has limited temporal robustness.

\item \textbf{Bottom-up keypoints + PAFs (OpenPose family~\cite{cao2019openpose}).}
Another option is to use bottom-up keypoints with PAFs, keeping multiple high-score limb connections via PAF line integrals and applying ROI/corridor/direction heuristics. However, this approach is sensitive to thresholds and cross-person confusion. Duplicated limbs are often suppressed by NMS or split across different persons, and recall drops in low-texture regions or fast motions.

\item \textbf{Takeaway.}
Overall, methods that assume a \emph{single} clean instance tend to hide limb duplication. We therefore adopt a detector trained specifically on structural artifacts and further incorporate temporal consistency.
\end{itemize}

\paragraph{\texttt{(II) Bone length.}}
\begin{itemize}\itemsep 2pt
\item \textbf{Sequence-level SMPL.}
Using sequence-level SMPL with shared $\beta$ and a near-constant global scale makes bone lengths \emph{intrinsically constant}, which makes the representation blind to local stretching or segment-wise distortions.

\item \textbf{Per-frame $\beta_t$ or only global $s_t$.}
Allowing per-frame shape $\beta_t$ breaks this constancy but leads to noisy and unstable estimates due to monocular depth ambiguity and is computationally expensive. In contrast, freeing only a per-frame global scale $s_t$ uniformly scales the entire body and \emph{cannot reveal local} bone-length changes.

\item \textbf{Takeaway.}
We therefore adopt the skeleton fitting pipeline (\figureautorefname~\ref{fig:fitting_pipeline}) with \emph{time-varying scale} vectors and a sparse scale mixer, which explicitly tests the temporal stability of each segment and makes local inconsistencies observable.
\end{itemize}

\paragraph{\texttt{(III) Joint range of motion.}}
\begin{itemize}\itemsep 2pt
\item \textbf{Learned pose plausibility (discriminators/priors).}
A common approach is to use a learned pose prior that captures a \emph{pose-conditioned} feasible set of joint angles (e.g., Akhter \& Black's pose-conditioned joint-angle limits~\cite{akhter2015pose}) and then score poses by their prior likelihood. While this is powerful for 3D reconstruction, it creates three practical problems for our RoM metric:  
(i) the score is \emph{holistic} over a high-dimensional joint-angle space and thus \emph{not interpretable per DoF}, which breaks our per-DoF violation counting and our frequency–severity–persistence aggregation;  
(ii) the feasible region and parameterization are \emph{skeleton/rule specific} (rotation order, local frames), making calibration across rigs and mapping to our OpenSim DoFs nontrivial;  
(iii) limits estimated from specific datasets are \emph{data dependent} (the original paper notes that standard MoCap sets are insufficient to learn true limits in all poses), so cross-dataset and cross-generator calibration becomes fragile. For these reasons, we opted for explicit anatomical ranges with a small tolerance on OpenSim DoFs rather than a learned plausibility score~\cite{akhter2015pose}.

\item \textbf{Dataset heuristics or 2D-angle rules.}
We also experimented with simple thresholding on 2D joint angles or dataset-derived heuristics. These rules are \emph{skeleton- and view-dependent}: thresholds change with the keypoint convention and camera orientation; self-occlusion and foreshortening create spurious ``violations''; and retargeting to different rigs alters the numeric ranges. Such effects made the metric unstable across actions and models, so we discarded this option in favor of 3D, rig-consistent limits on OpenSim DoFs.

\item \textbf{Takeaway.}
Mapping to \emph{OpenSim} and checking per-DoF anatomical limits with a small tolerance yields stable and interpretable violation signals.
\end{itemize}

\paragraph{\texttt{(IV) Self collision.}}
\begin{itemize}\itemsep 2pt

\item \textbf{SMPL fitting with HMR-2.0~\cite{10378229}.}
In practice, the SMPL meshes predicted by HMR-2.0 frequently introduce \emph{method-induced} self-intersections that are unrelated to the actual difficulty of the motion. This is consistent with recent reports that common HMR baselines can exhibit self-intersections, and that collision-aware post-processing (e.g., CLOAF~\cite{davydov2024cloaf}) is required to remove them (CLOAF reports $0.0\%$ collision after their correction). Using HMR-2.0 inflated our collision counts and confounded the metric, which is why we avoided it for scoring.

\item \textbf{Image-based HMR.}
Image-based HMR often shows noticeable \emph{temporal jitter/flicker}: small pose fluctuations (jitter) cause spurious collisions, which directly harms our video-level aggregation. For stability, we therefore switched to a monocular video fitter~\cite{shen2024gvhmr} with better temporal coherence.

\item \textbf{SDF-based penetration tests.}
These methods require watertight fields, which are not only computationally heavier but also sensitive to skin and clothing geometry, making them less practical for large-scale evaluation.

\item \textbf{Takeaway.}
We decided on using a monocular fitter that leverages global trajectories (i.e., GVHMR~\cite{shen2024gvhmr}) and BVH-accelerated triangle tests with non-local adjacency filtering to obtain robust per-frame collision signals.
\end{itemize}

\paragraph{\texttt{(V) Kinematic extremes.}}
\begin{itemize}\itemsep 2pt
\item \textbf{Pixel/flow magnitude as speed.}
Confounded by camera motion and appearance changes; not body-specific nor physically meaningful.
\item \textbf{Joint-angle speeds only.}
Misses cases where segment \emph{linear} speed is implausible even if angles are moderate; also under-penalizes long-lever endpoints.
\item \textbf{Reference-trajectory matching.}
Requires task-specific ground truth and does not scale to open-ended prompts.
\item \textbf{Takeaway.} We compute both \emph{joint angular} and \emph{segment linear} velocities against natural human limits and aggregate with frequency–severity–persistence.
\end{itemize}

\paragraph{\texttt{(VI) Motion smoothness.}}
\begin{itemize}\itemsep 2pt
\item \textbf{SMPL Jitter Degree (jerk-based soft scores).}
Jerk/acceleration–based smoothness indices are meaningful but \emph{highly sensitive} to sampling rate, filtering choices, and measurement noise; moreover, they can \emph{conflate legitimate fast motions} (sprints, strikes, jumps) with reconstruction noise, and units/scale are not directly comparable across rigs without careful normalization. Because our goal is to flag \emph{hard} biomechanical implausibility, we use explicit angular-acceleration (and optional jerk-energy) thresholds grounded in biomechanics, followed by the same frequency–severity–persistence aggregation. \cite{balasubramanian2015analysis, roren2022assessing, singh2021motion}

\item \textbf{Optical-flow temporal gradients/warping error.}
Flow-based temporal errors mainly track \emph{appearance/camera changes} and are not skeleton-specific; they respond to texture flicker or camera motion even when body kinematics are plausible, and we observed weak correlation with perceived kinematic jerk. Hence we do not include them for smoothness scoring in our setting.
\item \textbf{Takeaway.} We thus chose \emph{hard} angular-acceleration (and optional jerk-energy) thresholds anchored in biomechanics literature, followed by applying our standard video-level aggregation.
\end{itemize}

\subsection{Additional Metrics Considered}\label{sec:additional_metrics}

Beyond the six metrics reported in the main paper, we also investigated several other biomechanical signals. For each, we describe the goal, the implementations we tried, and the reason it is not included in the final benchmark.

\paragraph{\texttt{(I) Foot--Ground Contact.}}
The goal is to detect stance-phase sliding (``foot skating'') and foot floating or penetration into the ground. We tried two implementations: (i) fitting SMPL to the person and reconstructing the ground as a point cloud, then testing SMPL--ground contact in 3D; in practice this required accurate scale and camera alignment (single-view SMPL has scale/translation ambiguity), and SMPL resolves fine foot geometry poorly. (ii) Replacing the SMPL mesh with depth back-projection: obtaining person/foot masks from the image, back-projecting the same-frame depth into point clouds for feet and ground, and testing 3D contact directly; this reduced alignment error but depended heavily on depth quality.

We exclude this metric because, across our prompts and models, genuine foot--ground violations were rare; most failures were dominated by unfaithful foot or ground appearance, where neither implementation is reliable. The resulting score distribution became quasi-binary (near 0 or near 100), with low discriminative power and high sensitivity to depth noise, and therefore contributed little to model ranking.

\paragraph{\texttt{(II) Mechanical Feasibility (Inverse Dynamics).}}
The goal is to check whether joint torques or powers exceed human capabilities. Classical inverse dynamics requires accurate external forces (e.g., from force plates), which are unavailable from monocular video. We therefore attempted a surrogate pipeline: (1) estimate the whole-body resultant force via $F{=}ma$ from kinematics; (2) distribute this force between the left and right feet using stance heuristics (e.g., 100\% to the stance foot in single support, linearly interpolated otherwise); (3) compute joint torques via inverse dynamics on an OpenSim skeleton.

We exclude this metric because the force-partition step introduces large errors in complex, nonperiodic motions (multi-contact, aerial phases, rapid transitions), yielding unstable and uninterpretable torque estimates. Without a reliable force-partition model, the resulting signal is unsuitable for benchmarking.

\paragraph{\texttt{(III) Motion Stability (CoM vs. Support Polygon).}}
The goal is to test whether the body's center-of-mass (CoM) projection lies within the support polygon. This criterion is meaningful for quasi-static balance and robotics control, but humans routinely perform intentionally unstable actions (running, jumping, cutting, falling) that appear in our prompts. The metric therefore confounds legitimate dynamic behavior with actual errors, and additionally depends on reliable contact detection, which is itself uncertain from single-view video. As a result, it did not separate good from bad generations in our setting and we do not include it.

\section{Metric Robustness Analysis} \label{sec:robustness_app}

This appendix provides the detailed protocols and per-model results for the three robustness analyses summarized in \sectionautorefname~\ref{sec:metric_robustness} of the main paper.

\subsection{Robustness to Pose Estimation Method}

For metrics that rely on SMPL sequences, we evaluate robustness by replacing the pose estimator and recomputing all metrics for comparison. We consider two alternative pipelines. First, we apply additional temporal optimization to MeTRAbs keypoints to remove outliers and refit the cleaned keypoints to SMPL skeletons. Second, we use another off-the-shelf pose estimator, PromptHMR~\cite{wang2025prompthmr}, to directly obtain SMPL sequences and compute the metrics as normal. Across both alternatives, we observe identical model rankings, confirming that our evaluation is robust to the choice of pose estimation method.

\subsection{Robustness to Tolerance Scales}

\begin{figure}[h]
    \centering
    \includegraphics[width=0.6\textwidth]{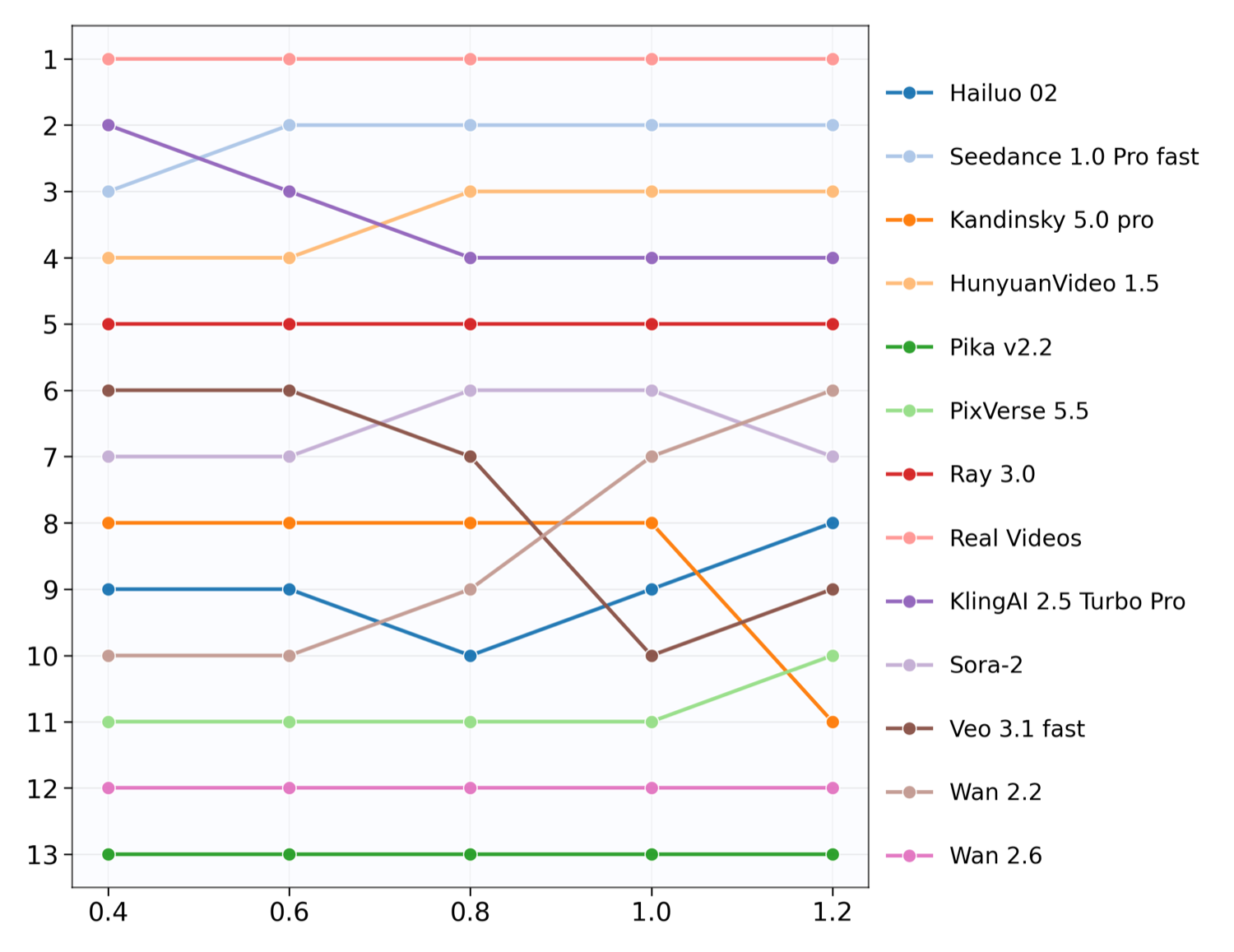}
    \caption{Model rankings (y-axis) across different tolerance scales (x-axis). Rankings remain consistent across the sweep, demonstrating the robustness of our metrics to tolerance choice.}
    \label{fig:scale_sweep}
\end{figure}

To accommodate natural estimation variability and motion diversity, we sweep the tolerance scales applied to anatomical limits (and analogously to bone-length, velocity, and acceleration thresholds) from strict to permissive. \figureautorefname~\ref{fig:scale_sweep} reports the resulting model rankings across the entire sweep; rankings remain consistent throughout, supporting the stability of our metrics with respect to tolerance choice.

\subsection{Robustness to Aggregation Hyperparameters}

\begin{figure}[h]
    \centering
    \includegraphics[width=0.99\textwidth]{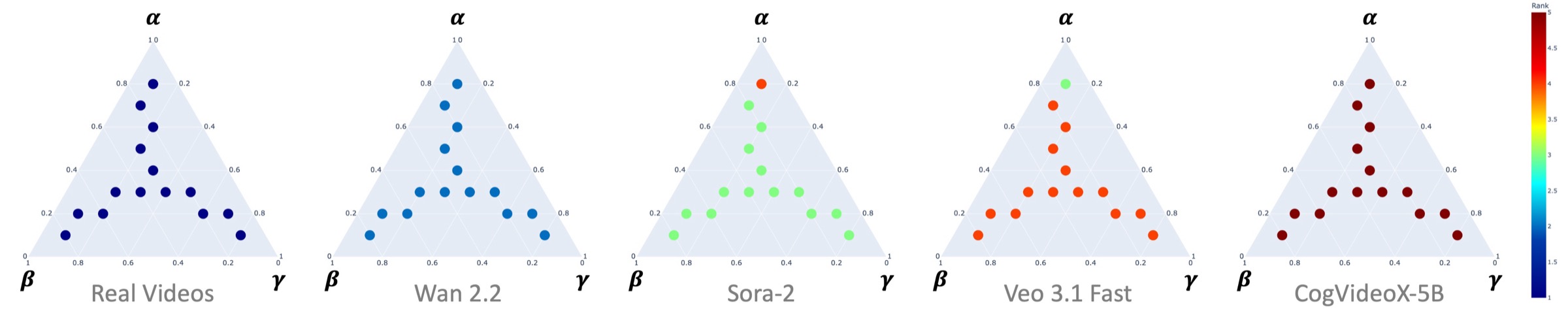}
    \caption{Ternary plots of model rankings under varying aggregation hyperparameters. Each point in a ternary diagram corresponds to one combination of $(\alpha, \beta, \gamma)$. For each model, the ranking remains consistent across different coordinates.}
    \label{fig:hyperparam_robust}
\end{figure}

The video-level score for each metric is a weighted combination of frequency $r$, severity $s$, and persistence $p$ with weights $(\alpha, \beta, \gamma)$, $\alpha+\beta+\gamma=1$. While many parameters in HumanScore are prescribed by biomechanics standards, $(\alpha, \beta, \gamma)$ are user-defined; we therefore perform an extensive grid search over the simplex and visualize the resulting model rankings as ternary plots in \figureautorefname~\ref{fig:hyperparam_robust}. Without loss of generality, we conduct this analysis on two open-source models (Wan 2.2~\cite{wan2025wan}, CogVideoX-5B~\cite{yang2024cogvideox}) and two proprietary models (Sora-2~\cite{openai_sora2_system_card_2025}, Veo 3.1 fast~\cite{googleveo31}). In all cases, the relative rankings of AI-generated and real videos remain consistent across the simplex, supporting the robustness of our method to aggregation-weight choice.

\subsection{\cradded{Reconstruction Stress Test}}\label{sec:reconstruction_stress_app}
\cradded{Gaussian noise on real-video poses ($\sigma=\mathrm{std}/(p_{95}-p_5)$ per DoF) gives scores of $96.2$, $96.4$, $95.8$, and $94.4$ at $\sigma=0$, $0.01$, $0.05$, and $0.1$. All exceed the strongest generator ($92.6$).}

\subsection{\cradded{Repeated Generation}}\label{sec:repeated_generation_app}
\cradded{The leaderboard uses one video per prompt and model. On 12 prompts (four per difficulty), three seeds give $92.0\pm1.1$ for Seedance, $90.2\pm1.4$ for Ray, and $87.7\pm2.0$ for HunyuanVideo (mean $\pm$ SD of seed-level means). Ray and HunyuanVideo exchange order, cautioning against strict rankings of nearby entries.}

\section{Statistical Results for HumanScore}\label{sec:stats_app}

To complement the per-tier averages reported in \tableautorefname~\ref{tab:model_compare}, we provide per-model dispersion statistics for each of the six HumanScore metrics on the 102-video AI subset (\tableautorefname~\ref{tab:per_model_stats}). \cradded{For each listed (metric, model) pair we report two quantities:} the within-model sample standard deviation (SD), which captures how widely a model's per-video scores spread across motions and intensities, and the standard error of the mean (SEM, $\mathrm{SD}/\sqrt{n}$ with $n{=}102$), which captures how precisely each leaderboard mean is pinned down by our 102 prompts. Read together, these two statistics make it easy to tell whether ranking gaps in \tableautorefname~\ref{tab:model_compare} reflect genuine differences or sampling noise: gaps that span several SEMs---which is the case between the top tier (Seedance, KlingAI) and the bottom tier (Sora-2, Kandinsky) across nearly every metric---\cradded{suggest broader performance differences, while nearby means require caution. These SEMs describe variation across prompts, not repeated generations. Under an independent-prompt normal approximation, a per-metric $95\%$ interval is $\bar S\pm1.96\,\mathrm{SEM}$; for example, Seedance\textquotesingle s extra-limb score is $94.2\pm2.6$. These are approximate intervals, not paired significance tests or intervals for the composite score.}

\begin{table*}[htb]
\caption{\textbf{Statistical results for HumanScore.} \cradded{For each listed (metric, model) pair,} we report the sample standard deviation (SD) and the standard error of the mean (SEM). A small SEM indicates that the leaderboard mean is well-estimated; a large SD indicates large within-model variance across motions and intensities.}
\renewcommand\tabcolsep{2pt}
\renewcommand\arraystretch{1.2}
\renewcommand{\thefootnote}{\fnsymbol{footnote}}
\setlength{\abovecaptionskip}{0.0cm}
\setlength{\belowcaptionskip}{-0.45cm}
\centering
\resizebox{\linewidth}{!}{%
\begin{tabular}{r |
>{\centering\arraybackslash}p{0.85cm}
>{\centering\arraybackslash}p{0.85cm}
>{\centering\arraybackslash}p{0.85cm}
>{\centering\arraybackslash}p{0.85cm}|
>{\centering\arraybackslash}p{0.85cm}
>{\centering\arraybackslash}p{0.85cm}
>{\centering\arraybackslash}p{0.85cm}
>{\centering\arraybackslash}p{0.85cm}|
>{\centering\arraybackslash}p{0.85cm}
>{\centering\arraybackslash}p{0.85cm}
>{\centering\arraybackslash}p{0.85cm}
>{\centering\arraybackslash}p{0.85cm}}
\toprule
\multicolumn{1}{c|}{\multirow{3}{*}{Models}} &
  \multicolumn{4}{c|}{\textbf{Anatomy Correctness}} &
  \multicolumn{4}{c|}{\textbf{Kinematic Correctness}} &
  \multicolumn{4}{c}{\textbf{\cradded{Kinetic Plausibility}}}
   \\
\cmidrule(lr){2-5} \cmidrule(lr){6-9} \cmidrule(lr){10-13}
 & \multicolumn{2}{c}{\texttt{(I)}} & \multicolumn{2}{c|}{\texttt{(II)}}
 & \multicolumn{2}{c}{\texttt{(III)}} & \multicolumn{2}{c|}{\texttt{(IV)}}
 & \multicolumn{2}{c}{\texttt{(V)}} & \multicolumn{2}{c}{\texttt{(VI)}} \\
\cmidrule(lr){2-3} \cmidrule(lr){4-5} \cmidrule(lr){6-7} \cmidrule(lr){8-9} \cmidrule(lr){10-11} \cmidrule(lr){12-13}
 & SD & SEM & SD & SEM & SD & SEM & SD & SEM & SD & SEM & SD & SEM \\
\hline
\hline
\rowcolor{row_color}
\multicolumn{13}{l}{\emph{Proprietary models}}\\
\includegraphics[height=0.75em]{sec/images/icons/medal1.png} Seedance 1.0 Pro fast~\cite{gao2025seedance} & $13.41$ & $1.33$ & $\phantom{0}6.91$ & $0.68$ & $22.64$ & $2.24$ & $10.77$ & $1.07$ & $12.79$ & $1.27$ & $10.60$ & $1.05$ \\
\includegraphics[height=0.75em]{sec/images/icons/medal2.png} KlingAI 2.5 Turbo Pro~\cite{kling25turbo} & $18.86$ & $1.87$ & $\phantom{0}7.99$ & $0.79$ & $25.46$ & $2.52$ & $11.18$ & $1.11$ & $17.91$ & $1.77$ & $18.18$ & $1.80$ \\
\includegraphics[height=0.75em]{sec/images/icons/medal3.png} Ray 3.0~\cite{ray30} & $23.54$ & $2.33$ & $\phantom{0}6.12$ & $0.61$ & $28.99$ & $2.87$ & $13.41$ & $1.33$ & $14.58$ & $1.44$ & $\phantom{0}7.94$ & $0.79$ \\
PixVerse 5.5~\cite{pixverse55} & $23.73$ & $2.35$ & $\phantom{0}7.76$ & $0.77$ & $30.41$ & $3.01$ & $12.44$ & $1.23$ & $13.64$ & $1.35$ & $10.79$ & $1.07$ \\
Hailuo 02~\cite{hailuo02} & $24.64$ & $2.44$ & $\phantom{0}8.01$ & $0.79$ & $30.90$ & $3.06$ & $21.13$ & $2.09$ & $24.30$ & $2.41$ & $15.83$ & $1.57$ \\
Veo 3.1 fast~\cite{googleveo31} & $26.46$ & $2.62$ & $12.39$ & $1.23$ & $27.24$ & $2.70$ & $10.53$ & $1.04$ & $17.82$ & $1.76$ & $19.60$ & $1.94$ \\
Wan 2.6~\cite{wan26} & $22.98$ & $2.28$ & $\phantom{0}6.71$ & $0.66$ & $32.01$ & $3.17$ & $11.02$ & $1.09$ & $30.58$ & $3.03$ & $25.03$ & $2.48$ \\
Pika v2.2~\cite{pika22} & $18.47$ & $1.83$ & $\phantom{0}7.20$ & $0.71$ & $25.15$ & $2.49$ & $14.95$ & $1.48$ & $35.57$ & $3.52$ & $34.63$ & $3.43$ \\
Sora-2~\cite{openai_sora2_system_card_2025} & $15.69$ & $1.55$ & $14.41$ & $1.43$ & $29.17$ & $2.89$ & $13.23$ & $1.31$ & $39.46$ & $3.91$ & $41.66$ & $4.13$ \\
\hline
\hline
\rowcolor{row_color}
\multicolumn{13}{l}{\emph{Open-sourced models}}\\
\includegraphics[height=0.75em]{sec/images/icons/medal1.png} HunyuanVideo 1.5~\cite{wu2025hunyuanvideo} & $12.04$ & $1.19$ & $\phantom{0}3.91$ & $0.39$ & $28.26$ & $2.80$ & $20.54$ & $2.03$ & $27.64$ & $2.74$ & $27.63$ & $2.74$ \\
Wan 2.2~\cite{wan2025wan} & $\phantom{0}8.42$ & $0.83$ & $\phantom{0}8.42$ & $0.83$ & $29.75$ & $2.95$ & $16.05$ & $1.59$ & $33.80$ & $3.35$ & $32.94$ & $3.26$ \\
Kandinsky 5.0 pro~\cite{kandinsky50} & $25.44$ & $2.52$ & $\phantom{0}6.63$ & $0.66$ & $33.06$ & $3.27$ & $11.73$ & $1.16$ & $36.09$ & $3.57$ & $37.69$ & $3.73$ \\
\bottomrule
\end{tabular}%
}
\label{tab:per_model_stats}
\end{table*}

\section{Details on Video Generators}

As mentioned in \sectionautorefname~\ref{sec:experiments}, we evaluate thirteen representative modern video generators in this work. Below we briefly describe each model, grouped into open-source and proprietary systems.

\subsection{Open-source Models}
\begin{itemize}
    \item \textbf{HunyuanVideo 1.5}~\cite{wu2025hunyuanvideo}.
    HunyuanVideo 1.5 is Tencent's open video generation model family described in a technical report, with upgrades over earlier HunyuanVideo releases in prompt following, motion quality, and high-resolution video synthesis.
    It is positioned as a general-purpose text-to-video generator for longer and more visually coherent clips, with improved controllability and stronger video quality in practical generation settings.

    \item \textbf{Kandinsky 5.0 Pro}~\cite{kandinsky50}.
    Kandinsky 5.0 is a family of foundation models for image and video generation that includes both open research checkpoints and production-oriented variants.
    The Pro configuration used in our leaderboard emphasizes higher-quality video generation with stronger text adherence and more stable spatiotemporal synthesis than lightweight tiers.

    \item \textbf{Wan 2.2}~\cite{wan2025wan}. 
    Wan 2.2 is an open-source family of video generation models released under the Apache~2.0 license, supporting both 480p and 720p video generation.
    The models adopt a Mixture-of-Experts (MoE) video architecture with an emphasis on ``cinematic'' control, and expose reproducible inference pipelines for text-to-video and image-conditioned generation at 720p resolution.

    \item \textbf{CogVideoX-5B}~\cite{yang2024cogvideox}. 
    CogVideoX-5B is the 5B-parameter variant in the CogVideoX series, a diffusion-transformer text-to-video model designed to produce longer and more temporally consistent clips. 
    It employs a 3D causal variational autoencoder to compress video sequences in space and time, reducing computational cost while mitigating flickering and improving spatiotemporal coherence in the generated videos.
\end{itemize}

\subsection{Proprietary Models}
\begin{itemize}
    \item \textbf{Ray 3.0}~\cite{ray30}.
    Ray 3.0 is Luma AI's latest video generation model exposed through the Dream Machine product line.
    It targets higher-quality and more controllable cinematic video generation than prior Ray releases, with better motion realism, scene consistency, and prompt responsiveness for short-form creative workflows.

    \item \textbf{KlingAI 2.5 Turbo Pro}~\cite{kling25turbo}.
    Kling is a commercial text- and image-to-video model developed by Kuaishou.
    The 2.5 Turbo Pro tier is tuned for high-fidelity, cinematic outputs with strong prompt adherence and smooth motion.
    Cloud deployments typically expose configurable clip duration (short clips), output resolution (up to 1080p), aspect ratio (landscape, portrait, square), classifier-free guidance strength, and negative prompts.

    \item \textbf{Seedance 1.0 Pro Fast}~\cite{gao2025seedance}. 
    Seedance 1.0 is ByteDance's multi-shot video generation model that can synthesize 1080p videos from either text prompts or reference images, with a focus on improved semantic understanding, prompt following, and cinematic aesthetics. 
    In our experiments we use the ``Pro Fast'' configuration provided by our API vendor, which wraps Seedance 1.0 in a latency-optimized preset designed for smooth, stable human motion, multi-shot storytelling across several camera views, and diverse visual styles.

    \item \textbf{Sora-2}~\cite{openai_sora2_system_card_2025}. 
    Sora-2 is OpenAI's second-generation, proprietary video-and-audio generation model. 
    Building on the original Sora, it is designed to produce high-resolution videos with more accurate physics, sharper realism, synchronized audio, and stronger controllability over style and scene dynamics. 
    It targets complex, multi-actor scenes and long-horizon storytelling, aiming to follow detailed natural-language prompts with high fidelity while remaining grounded in realistic world dynamics.

    \item \textbf{Veo 3.1 Fast}~\cite{googleveo31}. 
    Veo 3.1 is Google's latest video generator integrated into the Gemini ecosystem; it can create short, high-quality clips with native audio from text or from a single image. 
    The ``Fast'' variant uses the same underlying Veo 3.1 engine but is tuned for rapid generation, trading a small amount of peak quality for significantly lower latency. 
    It is exposed as a production API for creative workflows, supporting short-form video creation with controllable camera motion and prompt-driven narrative structure.

    \item \textbf{Hailuo 02}~\cite{hailuo02}.
    Hailuo 02 is a commercial short-video generation model in the MiniMax/Hailuo ecosystem that synthesizes clips from text prompts or reference images, with a stated focus on cinematic style, character rendering, and motion-rich scenes.

    \item \textbf{PixVerse 5.5}~\cite{pixverse55}.
    PixVerse 5.5 is a commercial text-to-video and image-to-video model exposed through developer-facing APIs.
    The 5.5 release emphasizes controllability, quality presets, and prompt- and style-driven motion control for short-form video production.

    \item \textbf{Wan 2.6}~\cite{wan26}.
    Wan 2.6 is a newer commercial release in the WAN video generation series, exposed in Model Studio alongside WAN 2.5/2.6 prompting guides and hosted generation services.
    Relative to the open Wan 2.2 checkpoints, it is presented as a more production-oriented offering with stronger prompt following and improved end-user generation quality.

    \item \textbf{Pika v2.2}~\cite{pika22}.
    Pika v2.2 is a commercial text-to-video model aimed at fast prompt-to-video generation of stylized short clips.
    It improves visual fidelity and motion quality over earlier Pika releases and adds editing-oriented controls.
\end{itemize}

\section{Future Work} \label{sec:future_work}

Looking ahead, we expect this benchmark to serve as a useful indicator for evaluating generated human motion from both open-source and proprietary video generation models. Researchers can use our metrics to obtain reliable insights into the faithfulness of generated motions that may already appear visually convincing at a perceptual level. We also view this benchmark as an important step for the computer vision, generative modeling, and biomechanics communities: feedback from our evaluation suite may serve as a valuable signal for improving both human motion generation and motion estimation models.

\section{Broader Impacts}

HumanScore is a benchmark for the biomechanical plausibility of human motion in AI-generated videos. We outline the principal positive and negative impacts we anticipate.

\paragraph{Positive impacts.}
HumanScore provides a quantitative, interpretable target for improving human-motion fidelity in generative video models. Closing the gap between visually realistic and biomechanically realistic generation is relevant to a broad set of beneficial applications: rehabilitation and motor-learning visualization, accessibility tools (e.g., sign-language synthesis where joint feasibility matters), educational and training media, and ergonomic or animation assistance for non-experts. Because our metrics are decomposed into anatomy, kinematics, and kinetics tiers, model developers can identify and target specific weaknesses rather than chasing a single opaque quality score; this transparency makes failure modes easier to diagnose and report.

\paragraph{Negative impacts and mitigations.}
Improving the biomechanical fidelity of generated videos also makes them harder to distinguish from real footage and may therefore lower the cost of producing convincing deepfakes or manipulated content depicting real people. We do not view this as a reason to withhold the benchmark---pixel-level visual realism is already advanced, and every generator evaluated here is publicly available---but we note that the same diagnostic signal can be repurposed by detection and provenance tools to flag biomechanical artifacts in suspect content. We encourage researchers and platforms to adopt such defensive uses alongside generative ones, and we include real-versus-AI separation results (\sectionautorefname~\ref{sec:realorai}) precisely to surface this dual use.

\paragraph{Bias and representativeness.}
The biomechanical limits and tolerance margins used by HumanScore are drawn from adult, typical-population biomechanics literature. Atypical bodies---children, elite athletes operating near genuine human limits, populations with non-typical morphology, and individuals with disabilities---may legitimately fall outside these conservative bounds, in which case our metrics could under-credit their motion. We treat this as a calibration limitation rather than a statement about the population, and we view a more population-aware reference set as a valuable extension. Real-world videos were collected from publicly available platforms (YouTube, Bilibili, YouTube-8M~\cradded{\cite{bilibili,youtube8m}}) for academic evaluation only; HumanScore consumes metric outputs, not the videos themselves, and we do not redistribute the underlying video material.

\section{Declaration of LLM Usage}

In the preparation of this manuscript, we used large language models (LLMs) as writing assistants to help refine wording, improve readability, tighten paragraph structure, and check English grammar. All scientific contributions---research questions, benchmark design, motion-set curation, prompt design, metric design, experimental protocol, results, analyses, and conclusions---were conceived, executed, and verified by the human authors. LLM-generated text was reviewed, edited, and approved by the authors before inclusion. \cradded{The MLLM-baseline experiment separately reports evaluator outputs as empirical measurements (\sectionautorefname~\ref{sec:mllm_comparison}).}

We additionally used an LLM as a categorization aid during motion-set curation (\sectionautorefname~\ref{sec:motion_curation}): the LLM was prompted to group candidate motions into families (e.g., \textit{Gait and Footwork}, \textit{Gymnastics and Calisthenics}) and to verify breadth of coverage across upper-body, lower-body, full-body, flips/rotations, self-contact, and object-interaction categories. Every grouping and coverage decision returned by the LLM was reviewed by the authors, and the final 51-motion set is the product of human judgement informed by, but not delegated to, the LLM. No LLM was used to fit, run, or interpret any HumanScore metric.

\section{Full List of Prompts}
In this benchmark, we include 51 unique motion types, each paired with gentle and intense prompt variants, for a total of 102 prompts. The motions are balanced across three difficulty levels: 17 easy, 17 medium, and 17 hard. In \figureautorefname~\ref{fig:prompt_stats}, we show the motion names and the most frequent words appearing in the prompts. Below, we provide the full prompt list used to query video generators in \autoref{tab:kinetics700_prompts1}, \autoref{tab:kinetics700_prompts2}, \autoref{tab:kinetics700_prompts3}, \autoref{tab:kinetics700_prompts4}, \autoref{tab:kinetics700_prompts5}, and \autoref{tab:kinetics700_prompts6}.

\begingroup
\small
\setlength{\LTleft}{0pt}
\setlength{\LTright}{0pt}
\begin{longtable}{|p{2.9cm}|p{1.6cm}|p{6.3cm}|}
\caption{Full list of prompts from the 51-motion benchmark (page 1).}\label{tab:kinetics700_prompts1}\\
\hline
\textbf{Motion} & \textbf{Difficulty} & \textbf{Prompt} \\
\hline
\endfirsthead
\hline
\textbf{Motion} & \textbf{Difficulty} & \textbf{Prompt} \\
\hline
\endhead
\hline
\endfoot
\hline
\endlastfoot
walking & easy & A single, full-body person walking in an indoor studio with a neutral background. Gentle motion; small, even steps and slow arm swing at an easy cadence. Locked, static camera; the subject stays centered and unobstructed. \\
\hline
walking & easy & A single, full-body person walking in an indoor studio with a neutral background. Intense motion; long, driving strides with vigorous arm swing at a quick pace. Locked, static camera; the subject stays centered and unobstructed. \\
\hline
jogging & easy & A single, full-body person jogging in an indoor studio with a neutral background. Gentle motion; short strides and relaxed arm drive at a light pace. Locked, static camera; the subject stays centered and unobstructed. \\
\hline
jogging & easy & A single, full-body person jogging in an indoor studio with a neutral background. Intense motion; extended stride length and forceful arm drive at a fast cadence. Locked, static camera; the subject stays centered and unobstructed. \\
\hline
jumping jacks & medium & A single, full-body person performs jumping jacks in an indoor studio with a neutral background. Gentle motion; low-amplitude jacks with soft landings and controlled arm travel at an easy tempo. Locked, static camera; the subject stays centered and unobstructed. \\
\hline
jumping jacks & medium & A single, full-body person performs jumping jacks in an indoor studio with a neutral background. Intense motion; high-amplitude jacks at a fast cadence, arms fully overhead with springy rebounds. Locked, static camera; the subject stays centered and unobstructed. \\
\hline
clapping hands & easy & A single, full-body person performs clapping hands in an indoor studio with a neutral background. Gentle motion; light, measured claps at a slow tempo, elbows near the torso and minimal body sway. Locked, static camera; the subject stays centered and unobstructed. \\
\hline
clapping hands & easy & A single, full-body person performs clapping hands in an indoor studio with a neutral background. Intense motion; rapid, forceful claps with wide arm swings and brisk cadence, pronounced accents. Locked, static camera; the subject stays centered and unobstructed. \\
\hline
tiptoeing & easy & A single, full-body person tiptoeing in an indoor studio with a neutral background. Gentle motion; small-amplitude steps and slow arm swing at an easy cadence. Locked, static camera; the subject stays centered and unobstructed. \\
\hline
tiptoeing & easy & A single, full-body person tiptoeing in an indoor studio with a neutral background. Intense motion; long strides with vigorous arm drive at a fast cadence. Locked, static camera; the subject stays centered and unobstructed. \\
\hline
lunge & easy & A single, full-body person lunge in an indoor studio with a neutral background. Gentle motion; short step length with slow, controlled lowering and rise. Locked, static camera; the subject stays centered and unobstructed. \\
\hline
lunge & easy & A single, full-body person lunge in an indoor studio with a neutral background. Intense motion; long step length with deep range and quick, forceful drive to stand. Locked, static camera; the subject stays centered and unobstructed. \\
\hline
squat & easy & A single, full-body person squat in an indoor studio with a neutral background. Gentle motion; shallow depth and slow, controlled tempo. Locked, static camera; the subject stays centered and unobstructed. \\
\hline
squat & easy & A single, full-body person squat in an indoor studio with a neutral background. Intense motion; deep range to strong lockout at a brisk, powerful tempo. Locked, static camera; the subject stays centered and unobstructed. \\
\hline
stretching leg & easy & A single, full-body person stretching leg in an indoor studio with a neutral background. Gentle motion; slow, controlled reps with small excursion and steady tempo. Locked, static camera; the subject stays centered and unobstructed. \\
\hline
stretching leg & easy & A single, full-body person stretching leg in an indoor studio with a neutral background. Intense motion; large, full-range reps performed briskly with strong drive and quick return. Locked, static camera; the subject stays centered and unobstructed. \\
\hline
stretching arm & easy & A single, full-body person stretching arm in an indoor studio with a neutral background. Gentle motion; slow, controlled reps with small excursion and steady tempo. Locked, static camera; the subject stays centered and unobstructed. \\
\hline
\end{longtable}
\endgroup

\begingroup
\small
\setlength{\LTleft}{0pt}
\setlength{\LTright}{0pt}
\begin{longtable}{|p{2.9cm}|p{1.6cm}|p{6.3cm}|}
\caption{Full list of prompts from the 51-motion benchmark (page 2).}\label{tab:kinetics700_prompts2}\\
\hline
\textbf{Motion} & \textbf{Difficulty} & \textbf{Prompt} \\
\hline
\endfirsthead
\hline
\textbf{Motion} & \textbf{Difficulty} & \textbf{Prompt} \\
\hline
\endhead
\hline
\endfoot
\hline
\endlastfoot
stretching arm & easy & A single, full-body person stretching arm in an indoor studio with a neutral background. Intense motion; large, full-range reps performed briskly with strong drive and quick return. Locked, static camera; the subject stays centered and unobstructed. \\
\hline
front raises & easy & A single, full-body person front raises in an indoor studio with a neutral background. Gentle motion; slow, controlled reps with small excursion and steady tempo. Locked, static camera; the subject stays centered and unobstructed. \\
\hline
front raises & easy & A single, full-body person front raises in an indoor studio with a neutral background. Intense motion; large, full-range reps performed briskly with strong drive and quick return. Locked, static camera; the subject stays centered and unobstructed. \\
\hline
push up & easy & A single, full-body person push up in an indoor studio with a neutral background. Gentle motion; shallow range and slow tempo with steady alignment. Locked, static camera; the subject stays centered and unobstructed. \\
\hline
push up & easy & A single, full-body person push up in an indoor studio with a neutral background. Intense motion; chest-to-floor depth and strong lockout at a brisk tempo. Locked, static camera; the subject stays centered and unobstructed. \\
\hline
situp & easy & A single, full-body person situp in an indoor studio with a neutral background. Gentle motion; slow, controlled reps with small excursion and steady tempo. Locked, static camera; the subject stays centered and unobstructed. \\
\hline
situp & easy & A single, full-body person situp in an indoor studio with a neutral background. Intense motion; large, full-range reps performed briskly with strong drive and quick return. Locked, static camera; the subject stays centered and unobstructed. \\
\hline
exercising with ball & medium & A single, full-body person exercising with an exercise ball in an indoor studio with a neutral background. Gentle motion; slow, controlled reps with small excursion and steady tempo. Locked, static camera; the subject stays centered and unobstructed. \\
\hline
exercising with ball & medium & A single, full-body person exercising with an exercise ball in an indoor studio with a neutral background. Intense motion; large, full-range reps performed briskly with strong drive and quick return. Locked, static camera; the subject stays centered and unobstructed. \\
\hline
doing aerobics & easy & A single, full-body person doing aerobics in an indoor studio with a neutral background. Gentle motion; slow, controlled reps with small excursion and steady tempo. Locked, static camera; the subject stays centered and unobstructed. \\
\hline
doing aerobics & easy & A single, full-body person doing aerobics in an indoor studio with a neutral background. Intense motion; large, full-range reps performed briskly with strong drive and quick return. Locked, static camera; the subject stays centered and unobstructed. \\
\hline
tai chi & easy & A single, full-body person tai chi in an indoor studio with a neutral background. Gentle motion; slow, controlled reps with small excursion and steady tempo. Locked, static camera; the subject stays centered and unobstructed. \\
\hline
tai chi & easy & A single, full-body person tai chi in an indoor studio with a neutral background. Intense motion; large, full-range reps performed briskly with strong drive and quick return. Locked, static camera; the subject stays centered and unobstructed. \\
\hline
moon walking & medium & A single, full-body person moon walking in an indoor studio with a neutral background. Gentle motion; small, even steps and slow arm swing at an easy cadence. Locked, static camera; the subject stays centered and unobstructed. \\
\hline
moon walking & medium & A single, full-body person moon walking in an indoor studio with a neutral background. Intense motion; long, driving strides with vigorous arm swing at a quick pace. Locked, static camera; the subject stays centered and unobstructed. \\
\hline
sweeping floor & easy & A single, full-body person sweeping floor in an indoor studio with a neutral background. Gentle motion; short strokes and slow, careful pacing. Locked, static camera; the subject stays centered and unobstructed. \\
\hline
sweeping floor & easy & A single, full-body person sweeping floor in an indoor studio with a neutral background. Intense motion; long strokes at brisk tempo with assertive direction changes. Locked, static camera; the subject stays centered and unobstructed. \\
\hline
\end{longtable}
\endgroup

\begingroup
\small
\setlength{\LTleft}{0pt}
\setlength{\LTright}{0pt}
\begin{longtable}{|p{2.9cm}|p{1.6cm}|p{6.3cm}|}
\caption{Full list of prompts from the 51-motion benchmark (page 3).}\label{tab:kinetics700_prompts3}\\
\hline
\textbf{Motion} & \textbf{Difficulty} & \textbf{Prompt} \\
\hline
\endfirsthead
\hline
\textbf{Motion} & \textbf{Difficulty} & \textbf{Prompt} \\
\hline
\endhead
\hline
\endfoot
\hline
\endlastfoot
mowing lawn & easy & A single, full-body person mowing lawn in an indoor studio with a neutral background. Gentle motion; small reach and careful pacing with compact, slow movements. Locked, static camera; the subject stays centered and unobstructed. \\
\hline
mowing lawn & easy & A single, full-body person mowing lawn in an indoor studio with a neutral background. Intense motion; long, fast reaches and broad trunk rotation with assertive pacing. Locked, static camera; the subject stays centered and unobstructed. \\
\hline
slacklining & hard & A single, full-body person walking slowly on a slackline and keep balance in an indoor studio with a neutral background. Gentle motion; short reaches and slow corrections focused on quiet stabilization. Locked, static camera; the subject stays centered and unobstructed. \\
\hline
slacklining & hard & A single, full-body person walking slowly on a slackline and keep balance in an indoor studio with a neutral background. Intense motion; large reaches and quick corrections with dynamic stabilization. Locked, static camera; the subject stays centered and unobstructed. \\
\hline
playing ping pong & easy & A single, full-body person playing ping pong in an indoor studio with a neutral background. Gentle motion; compact swings and short recoveries at slow pace. Locked, static camera; the subject stays centered and unobstructed. \\
\hline
playing ping pong & easy & A single, full-body person playing ping pong in an indoor studio with a neutral background. Intense motion; full swings with fast recoveries and quick footwork. Locked, static camera; the subject stays centered and unobstructed. \\
\hline
playing badminton & medium & A single, full-body person playing badminton in an indoor studio with a neutral background. Gentle motion; compact swings and short recoveries at slow pace. Locked, static camera; the subject stays centered and unobstructed. \\
\hline
playing badminton & medium & A single, full-body person playing badminton in an indoor studio with a neutral background. Intense motion; full swings with fast recoveries and quick footwork. Locked, static camera; the subject stays centered and unobstructed. \\
\hline
playing tennis & medium & A single, full-body person playing tennis in an indoor studio with a neutral background. Gentle motion; compact swings and short recoveries at slow pace. Locked, static camera; the subject stays centered and unobstructed. \\
\hline
playing tennis & medium & A single, full-body person playing tennis in an indoor studio with a neutral background. Intense motion; full swings with fast recoveries and quick footwork. Locked, static camera; the subject stays centered and unobstructed. \\
\hline
playing volleyball & medium & A single, full-body person playing volleyball in an indoor studio with a neutral background. Gentle motion; small steps and light swings at slow pace. Locked, static camera; the subject stays centered and unobstructed. \\
\hline
playing volleyball & medium & A single, full-body person playing volleyball in an indoor studio with a neutral background. Intense motion; big steps and forceful swings with rapid footwork. Locked, static camera; the subject stays centered and unobstructed. \\
\hline
playing basketball & medium & A single, full-body person playing basketball in an indoor studio with a neutral background. Gentle motion; small steps and light swings at slow pace. Locked, static camera; the subject stays centered and unobstructed. \\
\hline
playing basketball & medium & A single, full-body person playing basketball in an indoor studio with a neutral background. Intense motion; big steps and forceful swings with rapid footwork. Locked, static camera; the subject stays centered and unobstructed. \\
\hline
playing frisbee & medium & A single, full-body person catching and throwing frisbee in an indoor studio with a neutral background. Gentle motion; compact swings and small steps with slow, controlled trunk rotation. Locked, static camera; the subject stays centered and unobstructed. \\
\hline
playing frisbee & medium & A single, full-body person catching and throwing frisbee in an indoor studio with a neutral background. Intense motion; full-arc swings, big steps, and rapid trunk rotation at high intensity. Locked, static camera; the subject stays centered and unobstructed. \\
\hline
playing soccer & medium & A single, full-body person passing and kicking soccer ball in an indoor studio with a neutral background. Gentle motion; low chamber and slow extension with controlled balance. Locked, static camera; the subject stays centered and unobstructed. \\
\hline
\end{longtable}
\endgroup

\begingroup
\small
\setlength{\LTleft}{0pt}
\setlength{\LTright}{0pt}
\begin{longtable}{|p{2.9cm}|p{1.6cm}|p{6.3cm}|}
\caption{Full list of prompts from the 51-motion benchmark (page 4).}\label{tab:kinetics700_prompts4}\\
\hline
\textbf{Motion} & \textbf{Difficulty} & \textbf{Prompt} \\
\hline
\endfirsthead
\hline
\textbf{Motion} & \textbf{Difficulty} & \textbf{Prompt} \\
\hline
\endhead
\hline
\endfoot
\hline
\endlastfoot
playing soccer & medium & A single, full-body person passing and kicking soccer ball in an indoor studio with a neutral background. Intense motion; high chamber and fast extension to end range with quick retraction. Locked, static camera; the subject stays centered and unobstructed. \\
\hline
playing squash & hard & A single, full-body person playing squash or racquetball in an indoor studio with a neutral background. Gentle motion; compact swings and short recoveries at slow pace. Locked, static camera; the subject stays centered and unobstructed. \\
\hline
playing squash & hard & A single, full-body person playing squash or racquetball in an indoor studio with a neutral background. Intense motion; full swings with fast recoveries and quick footwork. Locked, static camera; the subject stays centered and unobstructed. \\
\hline
hurdling & medium & A single, full-body person running fast and jumping over a waist-high hurdle with the lead leg high in an indoor studio with a neutral background. Gentle motion; smooth approach with controlled clearance, soft trail-leg recovery, and balanced landing. Locked, static camera; the subject stays centered and unobstructed. \\
\hline
hurdling & medium & A single, full-body person running fast and jumping over a waist-high hurdle with the lead leg high in an indoor studio with a neutral background. Intense motion; explosive takeoff with aggressive lead-leg drive, rapid trail-leg snap-down, and quick sprint out. Locked, static camera; the subject stays centered and unobstructed. \\
\hline
long jump & medium & A single, full-body person long jump in an indoor studio with a neutral background. Gentle motion; shortened run-up with a small hop and a quiet, balanced landing. Locked, static camera; the subject stays centered and unobstructed. \\
\hline
long jump & medium & A single, full-body person long jump in an indoor studio with a neutral background. Intense motion; powerful acceleration, explosive takeoff with driving knee, and a decisive landing. Locked, static camera; the subject stays centered and unobstructed. \\
\hline
golf driving & medium & A single, full-body person performing a golf drive in an indoor studio with a neutral background. Gentle motion; short backswings with relaxed tempo and a soft, controlled follow-through. Locked, static camera; the subject stays centered and unobstructed. \\
\hline
golf driving & medium & A single, full-body person performing a golf drive in an indoor studio with a neutral background. Intense motion; full backswing with explosive hip rotation and fast clubhead speed into a decisive follow-through. Locked, static camera; the subject stays centered and unobstructed. \\
\hline
skipping rope & medium & A single, full-body person jumping rope in an indoor studio with a neutral background. Gentle motion; small, rhythmic hops with compact wrist turns and steady cadence. Locked, static camera; the subject stays centered and unobstructed. \\
\hline
skipping rope & medium & A single, full-body person jumping rope in an indoor studio with a neutral background. Intense motion; rapid cadence with higher rebound and crisp wrist rotation. Locked, static camera; the subject stays centered and unobstructed. \\
\hline
pirouetting & hard & A single, full-body person lifting one foot and turning in place multiple times in an indoor studio with a neutral background. Gentle motion; soft, small arm phrases and short steps with slow body rotation. Locked, static camera; the subject stays centered and unobstructed. \\
\hline
pirouetting & hard & A single, full-body person lifting one foot and turning in place multiple times in an indoor studio with a neutral background. Intense motion; expansive arm lines, deep steps, and brisk torso rotation across the phrase. Locked, static camera; the subject stays centered and unobstructed. \\
\hline
swing dancing & hard & A single, full-body person doing swing steps with kicks, adding arm swings and small turns in an indoor studio with a neutral background. Gentle motion; soft, small arm phrases and short steps with slow body rotation. Locked, static camera; the subject stays centered and unobstructed. \\
\hline
swing dancing & hard & A single, full-body person doing swing steps with kicks, adding arm swings and small turns in an indoor studio with a neutral background. Intense motion; expansive arm lines, deep steps, and brisk torso rotation across the phrase. Locked, static camera; the subject stays centered and unobstructed. \\
\hline
side kick & medium & A single, full-body person performing a side kick in an indoor studio with a neutral background. Gentle motion; low chamber and slow extension with controlled balance. Locked, static camera; the subject stays centered and unobstructed. \\
\hline
side kick & medium & A single, full-body person performing a side kick in an indoor studio with a neutral background. Intense motion; high chamber and fast extension to end range with quick retraction. Locked, static camera; the subject stays centered and unobstructed. \\
\hline
\end{longtable}
\endgroup

\begingroup
\small
\setlength{\LTleft}{0pt}
\setlength{\LTright}{0pt}
\begin{longtable}{|p{2.9cm}|p{1.6cm}|p{6.3cm}|}
\caption{Full list of prompts from the 51-motion benchmark (page 5).}\label{tab:kinetics700_prompts5}\\
\hline
\textbf{Motion} & \textbf{Difficulty} & \textbf{Prompt} \\
\hline
\endfirsthead
\hline
\textbf{Motion} & \textbf{Difficulty} & \textbf{Prompt} \\
\hline
\endhead
\hline
\endfoot
\hline
\endlastfoot
drop kicking & medium & A single, full-body person drop kicking in an indoor studio with a neutral background. Gentle motion; low chamber and slow extension with controlled balance. Locked, static camera; the subject stays centered and unobstructed. \\
\hline
drop kicking & medium & A single, full-body person drop kicking in an indoor studio with a neutral background. Intense motion; high chamber and fast extension to end range with quick retraction. Locked, static camera; the subject stays centered and unobstructed. \\
\hline
roller skating & hard & A single, full-body person roller skating in an indoor studio with a neutral background. Gentle motion; short, slow glides with compact arm swing. Locked, static camera; the subject stays centered and unobstructed. \\
\hline
roller skating & hard & A single, full-body person roller skating in an indoor studio with a neutral background. Intense motion; long, powerful strokes with wide arm drive at speed. Locked, static camera; the subject stays centered and unobstructed. \\
\hline
skateboarding & hard & A single, full-body person skateboarding in an indoor studio with a neutral background. Gentle motion; short pushes and slow deck work with compact stance. Locked, static camera; the subject stays centered and unobstructed. \\
\hline
skateboarding & hard & A single, full-body person skateboarding in an indoor studio with a neutral background. Intense motion; long pushes and quick transitions with dynamic deck work. Locked, static camera; the subject stays centered and unobstructed. \\
\hline
bowling & easy & A single, full-body person performs bowling in an indoor studio with a neutral background. Gentle motion; short approach, low arm speed, and a controlled release. Locked, static camera; the subject stays centered and unobstructed. \\
\hline
bowling & easy & A single, full-body person performs bowling in an indoor studio with a neutral background. Intense motion; full four-step approach with a high backswing and a fast roll into a firm follow-through. Locked, static camera; the subject stays centered and unobstructed. \\
\hline
deadlifting & medium & A single, full-body person deadlifting in an indoor studio with a neutral background. Gentle motion; light hinge with small range and slow return. Locked, static camera; the subject stays centered and unobstructed. \\
\hline
deadlifting & medium & A single, full-body person deadlifting in an indoor studio with a neutral background. Intense motion; full hip extension to lockout with a quick, powerful drive. Locked, static camera; the subject stays centered and unobstructed. \\
\hline
standing on hands & hard & A single, full-body person standing on hands in an indoor studio with a neutral background. Gentle motion; steady handstand holds with small balance corrections. Locked, static camera; the subject stays centered and unobstructed. \\
\hline
standing on hands & hard & A single, full-body person standing on hands in an indoor studio with a neutral background. Intense motion; brisk kick-ups to handstand with quick weight shifts and controlled exits. Locked, static camera; the subject stays centered and unobstructed. \\
\hline
contorting & hard & A single, full-body person slowly bending the body into extreme shapes and holding each pose for a few seconds in an indoor studio with a neutral background. Gentle motion; shallow ranges with unhurried transitions and steady breathing. Locked, static camera; the subject stays centered and unobstructed. \\
\hline
contorting & hard & A single, full-body person slowly bending the body into extreme shapes and holding each pose for a few seconds in an indoor studio with a neutral background. Intense motion; deep ranges with quick but controlled pose changes and firm stabilization. Locked, static camera; the subject stays centered and unobstructed. \\
\hline
bending back & hard & A single, full-body person bending back in an indoor studio with a neutral background. Gentle motion; shallow backbends with slow entries and exits for quiet stabilization. Locked, static camera; the subject stays centered and unobstructed. \\
\hline
bending back & hard & A single, full-body person bending back in an indoor studio with a neutral background. Intense motion; deep backbends with brisk transitions and a steady, controlled finish. Locked, static camera; the subject stays centered and unobstructed. \\
\hline
yoga & hard & A single, full-body person practicing yoga in an indoor studio with a neutral background. Gentle motion; a slow-flow sequence with steady holds and smooth, quiet transitions. Locked, static camera; the subject stays centered and unobstructed. \\
\hline
\end{longtable}
\endgroup

\begingroup
\small
\setlength{\LTleft}{0pt}
\setlength{\LTright}{0pt}
\begin{longtable}{|p{2.9cm}|p{1.6cm}|p{6.3cm}|}
\caption{Full list of prompts from the 51-motion benchmark (page 6).}\label{tab:kinetics700_prompts6}\\
\hline
\textbf{Motion} & \textbf{Difficulty} & \textbf{Prompt} \\
\hline
\endfirsthead
\hline
\textbf{Motion} & \textbf{Difficulty} & \textbf{Prompt} \\
\hline
\endhead
\hline
\endfoot
\hline
\endlastfoot
yoga & hard & A single, full-body person practicing yoga in an indoor studio with a neutral background. Intense motion; large-amplitude poses with brisk, controlled transitions and strong lines. Locked, static camera; the subject stays centered and unobstructed. \\
\hline
dancing ballet & hard & A single, full-body person dancing ballet in an indoor studio with a neutral background. Gentle motion; soft, small arm phrases and short steps with slow body rotation. Locked, static camera; the subject stays centered and unobstructed. \\
\hline
dancing ballet & hard & A single, full-body person dancing ballet in an indoor studio with a neutral background. Intense motion; expansive arm lines, deep steps, and brisk torso rotation across the phrase. Locked, static camera; the subject stays centered and unobstructed. \\
\hline
figure skating & hard & A single, full-body person gliding on ice and performing small spins and crossovers. Gentle motion; short glides with light edge changes and slow rotations. Locked, static camera; the subject stays centered and unobstructed. \\
\hline
figure skating & hard & A single, full-body person gliding on ice and performing small spins and crossovers. Intense motion; faster strokes with stronger edge changes and quick spin entries. Locked, static camera; the subject stays centered and unobstructed. \\
\hline
high jump & hard & A single, full-body person high jump in an indoor studio with a neutral background. Gentle motion; short curved approach with a low, controlled takeoff and soft landing. Locked, static camera; the subject stays centered and unobstructed. \\
\hline
high jump & hard & A single, full-body person high jump in an indoor studio with a neutral background. Intense motion; fast curved approach with explosive takeoff and a pronounced arch over the bar path. Locked, static camera; the subject stays centered and unobstructed. \\
\hline
rock climbing & hard & A single, full-body person rock climbing in an indoor studio with a neutral background. Gentle motion; short reaches and slow pulls with controlled foot placements. Locked, static camera; the subject stays centered and unobstructed. \\
\hline
rock climbing & hard & A single, full-body person rock climbing in an indoor studio with a neutral background. Intense motion; long reaches and powerful pulls with rapid foot moves. Locked, static camera; the subject stays centered and unobstructed. \\
\hline
riding unicycle & hard & A single, full-body person riding unicycle in an indoor studio with a neutral background. Gentle motion; short reaches and slow corrections focused on quiet stabilization. Locked, static camera; the subject stays centered and unobstructed. \\
\hline
riding unicycle & hard & A single, full-body person riding unicycle in an indoor studio with a neutral background. Intense motion; large reaches and quick corrections with dynamic stabilization. Locked, static camera; the subject stays centered and unobstructed. \\
\hline
javelin throw & hard & A single, full-body person throwing a javelin in an indoor studio with a neutral background. Gentle motion; shortened approach with compact rotation and a soft, balanced release. Locked, static camera; the subject stays centered and unobstructed. \\
\hline
javelin throw & hard & A single, full-body person throwing a javelin in an indoor studio with a neutral background. Intense motion; rapid cross-steps, explosive hip--shoulder separation, and a decisive release. Locked, static camera; the subject stays centered and unobstructed. \\
\hline
swinging bat & medium & A single, full-body person swinging baseball bat and hitting baseball in an indoor studio with a neutral background. Gentle motion; small steps and light swings at slow pace. Locked, static camera; the subject stays centered and unobstructed. \\
\hline
swinging bat & medium & A single, full-body person swinging baseball bat and hitting baseball in an indoor studio with a neutral background. Intense motion; big steps and forceful swings with rapid footwork. Locked, static camera; the subject stays centered and unobstructed. \\
\hline
high kick & hard & A single, full-body person high kick in an indoor studio with a neutral background. Gentle motion; low chamber and slow extension with controlled balance. Locked, static camera; the subject stays centered and unobstructed. \\
\hline
high kick & hard & A single, full-body person high kick in an indoor studio with a neutral background. Intense motion; high chamber and fast extension to end range with quick retraction. Locked, static camera; the subject stays centered and unobstructed. \\
\hline
\end{longtable}
\endgroup

\newpage
\section*{NeurIPS Paper Checklist}

\begin{enumerate}

\item {\bf Claims}
    \item[] Question: Do the main claims made in the abstract and introduction accurately reflect the paper's contributions and scope?
    \item[] Answer: \answerYes{}
    \item[] Justification: The abstract and introduction state the paper's scope as a controlled benchmark for single-person, full-body human motion quality in generated videos, and the contributions match the prompt set, metric hierarchy, and model evaluation reported later. They also qualify the metrics as estimator-aware diagnostics rather than a perfect biomechanical oracle.

\item {\bf Limitations}
    \item[] Question: Does the paper discuss the limitations of the work performed by the authors?
    \item[] Answer: \answerYes{}
    \item[] Justification: The paper discusses limitations from monocular reconstruction, restricted single-person visible-body settings, and kinetic metrics that use velocity- and acceleration-based surrogates rather than direct force or torque estimates. It also reports robustness analyses for estimator choice, tolerance scales, and aggregation hyperparameters.

\item {\bf Theory assumptions and proofs}
    \item[] Question: For each theoretical result, does the paper provide the full set of assumptions and a complete (and correct) proof?
    \item[] Answer: \answerNA{}
    \item[] Justification: The paper does not present theoretical theorems or proofs. The mathematical expressions are metric definitions and aggregation formulas, with implementation details provided in the supplementary material.

    \item {\bf Experimental result reproducibility}
    \item[] Question: Does the paper fully disclose all the information needed to reproduce the main experimental results of the paper to the extent that it affects the main claims and/or conclusions of the paper (regardless of whether the code and data are provided or not)?
    \item[] Answer: \answerYes{}
    \item[] Justification: Our manuscript describes the benchmark prompts, metric definitions, datasets, model list, and evaluation metrics and their impelmentation details. 

\item {\bf Open access to data and code}
    \item[] Question: Does the paper provide open access to the data and code, with sufficient instructions to faithfully reproduce the main experimental results, as described in supplemental material?
    \item[] Answer: \answerYes{}
    \item[] Justification: An anonymized code repository is linked to the submission. 

\item {\bf Experimental setting/details}
    \item[] Question: Does the paper specify all the training and test details (e.g., data splits, hyperparameters, how they were chosen, type of optimizer) necessary to understand the results?
    \item[] Answer: \answerYes{}
    \item[] Justification: The paper specifies the 102-prompt benchmark, the evaluated models, the MoCap verification dataset, the real-video collection criteria, and the metric hyperparameters and fitting pipelines in the appendix. No training is performed by the proposed benchmark, so optimizer and training-split details are not applicable.

\item {\bf Experiment statistical significance}
    \item[] Question: Does the paper report error bars suitably and correctly defined or other appropriate information about the statistical significance of the experiments?
    \item[] Answer: \answerYes{}
    \item[] Justification: The paper reports leaderboard scores, human-preference correlations, and robustness sweeps. In supplementary materials, more statistical information is provided, including error bars and statistical significance tests.

\item {\bf Experiments compute resources}
    \item[] Question: For each experiment, does the paper provide sufficient information on the computer resources (type of compute workers, memory, time of execution) needed to reproduce the experiments?
    \item[] Answer: \answerYes{}
    \item[] Justification: The paper provides the platform used to call the video generation models. Given that the metric computation is not computationally intensive, the paper does not provide additional information on the compute resources needed to reproduce the experiments.
    
\item {\bf Code of ethics}
    \item[] Question: Does the research conducted in the paper conform, in every respect, with the NeurIPS Code of Ethics \url{https://neurips.cc/public/EthicsGuidelines}?
    \item[] Answer: \answerYes{}
    \item[] Justification: The research evaluates generated videos, public or cited datasets, and an anonymized human preference study, and the manuscript preserves submission anonymity. We are not aware of any aspect of the work that requires deviation from the NeurIPS Code of Ethics.

\item {\bf Broader impacts}
    \item[] Question: Does the paper discuss both potential positive societal impacts and negative societal impacts of the work performed?
    \item[] Answer: \answerYes{}
    \item[] Justification: There is a section in the supplementary material that discusses the potential negative societal impacts of the work.
    
\item {\bf Safeguards}
    \item[] Question: Does the paper describe safeguards that have been put in place for responsible release of data or models that have a high risk for misuse (e.g., pre-trained language models, image generators, or scraped datasets)?
    \item[] Answer: \answerNA{}
    \item[] Justification: The paper does not release a high-risk generative model or similarly dual-use model artifact. The benchmark assets are prompts, metric definitions, and evaluation results rather than a system that directly enables generation or misuse.

\item {\bf Licenses for existing assets}
    \item[] Question: Are the creators or original owners of assets (e.g., code, data, models), used in the paper, properly credited and are the license and terms of use explicitly mentioned and properly respected?
    \item[] Answer: \answerNo{}
    \item[] Justification: The paper cites the main datasets, tools, and video generators used, but it does not yet explicitly list licenses or terms of use for every existing asset and platform. This is especially relevant for datasets, pretrained fitting tools, proprietary generators, and public-video sources.

\item {\bf New assets}
    \item[] Question: Are new assets introduced in the paper well documented and is the documentation provided alongside the assets?
    \item[] Answer: \answerYes{}
    \item[] Justification: The paper introduces a new benchmark prompt suite and documents its construction, usage assumptions, and full prompt list in the supplementary material. The benchmark metrics are also documented with detailed formulas and implementation choices.

\item {\bf Crowdsourcing and research with human subjects}
    \item[] Question: For crowdsourcing experiments and research with human subjects, does the paper include the full text of instructions given to participants and screenshots, if applicable, as well as details about compensation (if any)? 
    \item[] Answer: \answerNo{}
    \item[] Justification: The paper reports a human preference study with approximately 1200 responses, but it does not include the full participant instructions, screenshots, compensation details, or full annotation protocol. These details should be added to the supplementary material.

\item {\bf Institutional review board (IRB) approvals or equivalent for research with human subjects}
    \item[] Question: Does the paper describe potential risks incurred by study participants, whether such risks were disclosed to the subjects, and whether Institutional Review Board (IRB) approvals (or an equivalent approval/review based on the requirements of your country or institution) were obtained?
    \item[] Answer: \answerNA{} 
    \item[] Justification: The manuscript includes a human preference study but does not describe participant risks, risk disclosure, or IRB/equivalent review status. These details should be reported in an anonymized form consistent with the relevant institutional requirements.

\item {\bf Declaration of LLM usage}
    \item[] Question: Does the paper describe the usage of LLMs if it is an important, original, or non-standard component of the core methods in this research? Note that if the LLM is used only for writing, editing, or formatting purposes and does \emph{not} impact the core methodology, scientific rigor, or originality of the research, declaration is not required.
    \item[] Answer: \answerYes{}
    \item[] Justification: The paper declares use of an LLM as part of the motion-set curation process, where candidate actions are grouped into motion families and checked for coverage. \cradded{It also reports MLLM-based evaluation baselines in \sectionautorefname~\ref{sec:mllm_comparison}.} This usage affects benchmark construction and is therefore described in the method section. In the supplementary material, we provide more details on the LLM usage to the paper writing.

\end{enumerate}

\end{document}